\documentclass[11pt]{article}

\usepackage[final]{acl}

\usepackage{times}
\usepackage{latexsym}

\usepackage[T1]{fontenc}

\usepackage[utf8]{inputenc}

\usepackage{microtype}

\usepackage{inconsolata}

\usepackage{graphicx}
\usepackage{subcaption}
\usepackage{booktabs}
\usepackage{hyperref}
\usepackage{multirow} 
\usepackage{array} 
\usepackage{amssymb} 
\usepackage[table]{xcolor}
\usepackage{rotating} 
\usepackage{wrapfig}
\usepackage[dvipsnames]{xcolor}
\usepackage{algorithm}
\usepackage{algorithmic}
\usepackage{enumitem}
\usepackage{amsmath}
\usepackage{amssymb}
\usepackage{mathtools}
\usepackage{amsthm}
\usepackage[textsize=tiny]{todonotes}
\usepackage[capitalize,noabbrev]{cleveref}
\usepackage{diagbox}

\usepackage{amsthm}

\newtheorem{theorem}{Theorem}

\title{Mitigating Model Merging Conflicts via Knowledge-Aware Task Vector Purification}

\author{
 \textbf{Bang An\textsuperscript{1}},\quad
 \textbf{Yibo Yang\textsuperscript{1}}\thanks{Corresponding author: Yibo Yang},\quad
 \textbf{Philip Torr\textsuperscript{2}},\quad
 \textbf{Bernard Ghanem\textsuperscript{1}}
\\
\\
 \textsuperscript{1}King Abdullah University of Science and Technology (KAUST),
 \textsuperscript{2}University of Oxford
\\
}

\begin{document}
\maketitle
\begin{abstract}
Model merging aims to integrate task-specific abilities from individually fine-tuned models into a single model without extra training.
In model merging methods, the task vector has become a fundamental building block, as it can encapsulate the information from fine-tuning. 
However, the merged model often suffers from performance degradation due to conflicts caused by task-irrelevant redundancy in task vectors. 
To address these challenges, we propose \textbf{P}urified T\textbf{A}sk \textbf{VE}ctors (PAVE) in knowledge-aware subspace. 
Concretely, we sample training examples from each task, and feed them into their fine-tuned models to acquire covariance matrices before linear layers.
We then perform a context-oriented singular value decomposition, which accentuates the weight components most relevant to the target knowledge.
As a result, we split fine-tuned model weights into task-relevant and redundant components in the knowledge-aware subspace, and purify the task vector by pruning redundant components.  
To induce fair pruning efforts across models, we further introduce a spectral rank allocation strategy by optimizing a normalized activated pruning error. 
The task vector purification by our method as a plug-and-play scheme is applicable across various task vector-based merging methods to improve their performance.
In experiments, we demonstrate the effectiveness of PAVE across diverse merging methods, tasks, and model architectures.\footnote{Accepted by The 2026 Conference on Empirical Methods in Natural Language Processing (EMNLP 2026). Our code and models are released at: \texttt{\url{https://github.com/ABgit111/purified_task_vector}}.}
\end{abstract}

\section{Introduction}
Transformer-based models, including large language models (LLMs) \citep{devlin2019bert,liu2019roberta,he2020deberta,radford2019language,touvron2023llama,achiam2023gpt} and vision models \citep{dosovitskiy2020image, radford2021learning} have found widespread applications across various domains. 
Supervised fine-tuning \citep{radford2018improving, hu2022lora} is a common technique to enhance downstream task performance. 
Despite the effectiveness of multi-task learning (MTL) in acquiring versatile abilities across multiple domains \citep{caruana1997multitask, zhang2018overview, vandenhende2021multi}, 
its applicability will be limited when data access is restricted or computational resources are limited.
To address this challenge, model merging has been proposed, aiming to construct a single model by combining parameters from domain-specific models, without relying on any training or fine-tuning efforts.

Among merging methods, Task Arithmetic \citep{ilharco2022editing} has achieved great success by introducing the task vector, defined as the difference between the fine-tuned model weights for a certain task and the base model. 
Task vectors capture the update direction of each task, enabling efficient composition of knowledge from multiple fine-tuned models while preserving their distinctive features. As a result, they have become a fundamental building block in subsequent studies \citep{ilharco2022editing, cheng2025whoever, du2024parameter, huang2024emr}.
However, plain task vectors often contain substantial redundancy, 
as not all parameter shifts during fine-tuning contribute to task-specific capabilities \citep{yu2024language}.
The redundant or noisy components may interfere with the valuable components from other task vectors during merging. 
As a result, performing model merging based on such task vectors still encounters conflicts and yields suboptimal performance \citep{yadav2023ties}.

\begin{figure*}
\begin{center}
\includegraphics[width=0.9\linewidth]{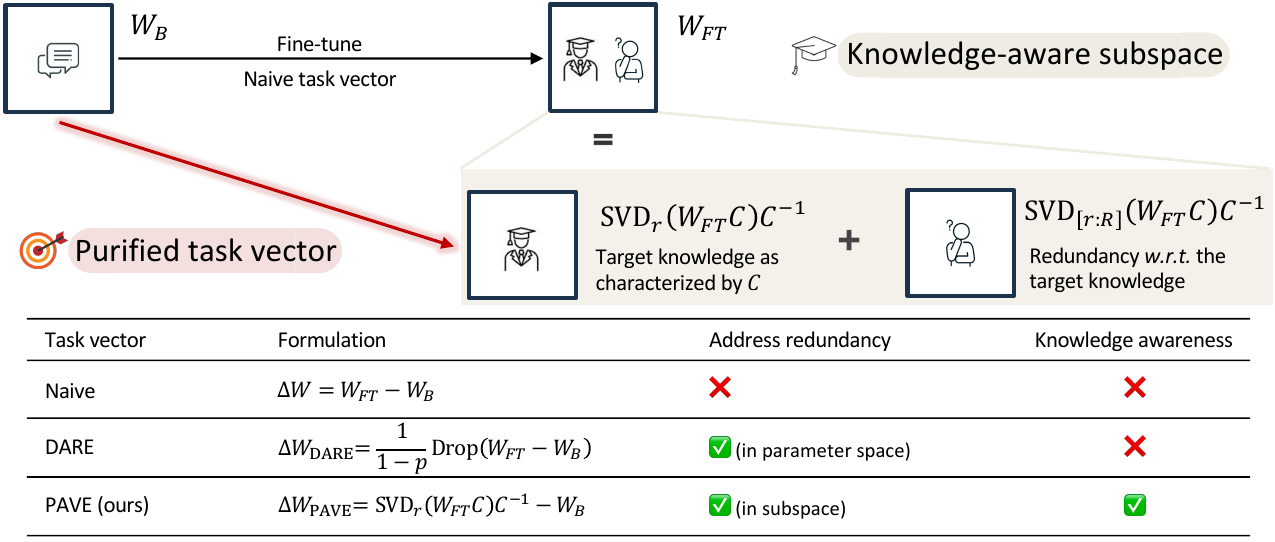} 
\end{center}
\caption{
An illustration of our method and a comparison between PAVE and prior task vectors. 
DARE addresses redundancy by randomly dropping elements in the parameter space without knowledge-awareness.
PAVE refines task vectors by identifying the target knowledge-relevant components within knowledge-aware subspace and removing the redundant components.}
\label{fig1}
\end{figure*}

To address the redundancy generally in task vectors, plug-in methods have been proposed. DARE \citep{yu2024language} proposes to randomly mask task vectors by Dropout \citep{srivastava2014dropout} to induce sparsity and reduce conflict. DOGE-Merging \citep{wei2025modeling} and AWD-Merging \citep{xiong2024multi} utilize plain SVD to evaluate components across tasks.
Despite improved performance, we point out that: 
(1) Random masks lack knowledge-awareness and may fail to accurately locate the parameter elements most relevant to a specific task; 
(2) Task-relevant and redundant elements are not necessarily separable in the parameter space, and actually task-specific ability lies in the subspaces of parameters \citep{raghu2017svcca,li2018measuring,hu2022lora}, which means that
masking in the parameter space may not effectively disentangle abilities and resolve conflicts.
(3) Plain SVD is also not knowledge-aware and component analysis based on it can be fragile and inaccurate; see \cref{subsec:task-aware-decomp}.
Therefore, how to purify the task vectors to make them precisely align with the task abilities is still a fundamental challenge in model merging.


To this end, we propose \textbf{p}urified t\textbf{a}sk \textbf{ve}ctors in knowledge-aware subspace (PAVE). 
PAVE builds on the context-oriented singular value decomposition, introduced by CorDA \cite{yang2024corda}. Specifically, given multiple fine-tuned models on different tasks, we randomly sample a small number of training samples for each task, and feed them into the corresponding fine-tuned model.
After obtaining the activation of each layer $X$, we calculate the covariance matrix $C=XX^T$ and then perform the context-oriented decomposition by applying SVD to the product of the fine-tuned weight matrix $W_{\text{FT}}$ and the covariance matrix $C$ from its input.
By doing so, the decomposed subspace is spanned by singular vectors that are ordered by their contributions to the task associated with $C$.
It has been observed that the first several components account for the majority of the learned task-relevant knowledge. 
Therefore, we filter out the redundant and noisy directions by retaining only the largest $r$ singular values and their singular vectors. 
We then post-multiply by $C^{-1}$ to reconstruct the fine-tuned model weights, which yields our purified task vector formulated as $\Delta W_{\text{PAVE}} = \text{SVD}_{r}(W_{\text{FT}}C)C^{-1} - W_{\text{B}}$.
The illustration of our method and its comparison with existing task vector constructions are shown in Figure~\ref{fig1}.


To induce fair pruning efforts across models, we further introduce a spectral rank allocation strategy that 
allocates the number of preserved ranks based on their spectral distribution.
It aims to minimize the summation of a normalized activated pruning error from each individual model. 
This objective can be efficiently optimized using a greedy algorithm that iteratively excludes components with the smallest normalized singular value.
It is noteworthy that PAVE as a plug-and-play method can be integrated on a wide range of task vector-based merging methods to improve their performance. 
We evaluate the performance of PAVE in combination with several popular merging strategies, including Task Arithmetic~\citep{ilharco2022editing}, Ties-Merging~\citep{yadav2023ties}, EMR-Merging~\citep{huang2024emr}, and WUDI-Merging~\citep{cheng2025whoever}. 
Experiments are conducted on the GLUE benchmark with two widely used language models, RoBERTa~\citep{liu2019roberta} and DeBERTa~\citep{he2020deberta}, and also on generation tasks, Math and Coding, with LLaMA-2-7B~\citep{touvron2023llama}. 
Additionally, we apply our approach to ViT-B/16, ViT-B/32, and ViT-L/14~\citep{dosovitskiy2020image} on 8 vision tasks. 

The contributions of this study are summarized as follows:
\begin{itemize}
    \item We propose PAVE, a plug-and-play method that applies knowledge-aware decomposition to task-vector purification, reducing task-irrelevant redundancy and improving modern model merging methods.
    \item We design a spectral rank allocation strategy to further refine the redundancy pruning process and improve performance.
    \item We demonstrate the effectiveness of PAVE with a wide range of merging methods, tasks, and model architectures. Notably, PAVE improves the performance of EMR-Merging on GLUE with RoBERTa from 80.18\% to 84.28\%, which is close to the average result of 8 individual models 85.55\%. 
\end{itemize}

\section{Related Work}
\textbf{Model Merging.} 
Model merging has emerged as an active area due to its great potential in integrating the abilities from multiple specialized models into a single model \citep{wortsman2022model, yadav2023ties, yang2023adamerging, xiao2023lm, lu2024twin}. Existing approaches include optimization-based methods~\citep{matena2022merging, jin2022dataless, wang2024localizing}, test-time adaptation methods~\citep{yang2023adamerging, yang2024representation, tang2023concrete}, and task vector-based methods~\citep{ilharco2022editing, wei2025modeling,xiong2024multi, sun2025task}. 
In particular, task vector has become an influential tool, as it encodes the essential information obtained during fine-tuning and can be effectively injected to a merged model using appropriate techniques. 
However, during the merging process, evidence has shown that there is noise and redundancy in the plain task vectors, which can cause conflicts and performance degradation. 
To address this issue, 
Ties-Merging~\citep{yadav2023ties} identifies redundancy in model parameters and proposes a trimming and sign-selection scheme to resolve conflicts. EMR-Merging~\citep{huang2024emr} integrates task components through electing, masking, and rescaling to reduce divergence between models. DARE~\citep{yu2024language}, as a plug-and-play method, resolves conflict issues by introducing random Dropout to the task vectors. These methods mainly reduce redundancy by removing elements from task vectors and can therefore be viewed as sparsity-based approaches.
In contrast, our method purifies task vectors in a knowledge-aware subspace by performing context-oriented decomposition and pruning task-irrelevant weight components.

Recent work has also explored decomposition-based analysis of task vectors. For example, Twin-Merging~\citep{lu2024twin}, TSV-Merging~\citep{gargiulo2025task}, DOGE-Merging~\citep{wei2025modeling}, and ISO-Merging~\citep{marczak2025no} apply plain SVD to task vectors for pruning. However, as we pointed out in \cref{subsec:task-aware-decomp}, the plain SVD is task-agnostic, this may cause inaccurate knowledge localization and suffer from substantial knowledge loss during rank pruning.  In comparison, our proposed method performs task vector purification in knowledge-aware subspaces, which helps identify and retain the components that are truly relevant to each task, leading to smoother and more effective merging.



\textbf{Low-Rank Decomposition for LLMs.} 
Low-rank decomposition based on SVD has been widely adopted in LLMs across multiple applications. For example, in low-rank adaptation \citep{hu2022lora,liu2024dora,zhang2023adalora,yang2024corda,yang2025dynamic}, SVD is often leveraged to create lightweight low-rank adapters for improving fine-tuning performance \citep{meng2024pissa} or preserving task-specific abilities \citep{yang2024corda, liang2025thanora}. 
In model compression, various methods have been proposed to prune model parameters based on the importance provided by singular values \citep{yuan2023asvd,wang2024svd,hsu2022language,li2025optimizing}.
Similar techniques have also been developed for quantization \citep{li2024loftq,li2024svdquant}. 

\section{Methodology}
\subsection{Preliminaries}
Given $\{\mathcal{M}\{W_1^l\}_{l=1}^L,\cdots, \mathcal{M}\{W_K^l\}_{l=1}^L\}$ that are models fine-tuned on $K$ different tasks $\{T_1, T_2, \cdots, T_K\}$ from the same base model $\mathcal{M}\{W_{\text{B}}^l\}_{l=1}^L$, where $L$ represents the number of layers, 
model merging aims to produce a single model that simultaneously achieves good performance on these tasks without extra training or fine-tuning.  
The merged model $\mathcal{M}\{W_{\text{M}}^l\}_{l=1}^L$ can be given by: 
\begin{align*}
   & \mathcal{M}\{W_{\text{M}}^l\}_{l=1}^L \\
   = & \phi(\mathcal{M}\{W_1^l\}_{l=1}^L,\cdots, \mathcal{M}\{W_K^l\}_{l=1}^L, \mathcal{M}\{W_{\text{B}}^l\}_{l=1}^L),
\end{align*}
where $\phi$ refers to a merging method.  

Task vector captures the task-specific changes introduced during fine-tuning and has become a common and essential tool in modern model merging methods. 
It is typically defined as the element-wise difference between a fine-tuned model and the base model,
\begin{equation*}
    \Delta W = W_{\text{FT}} - W_{\text{B}}.
\end{equation*}

However, recent studies have shown that naively calculating the difference between $W_{\text{FT}}$ and $W_{\text{B}}$ cannot precisely represent the shifting direction of a certain task in the parameter space, as 
the weight change caused by fine-tuning contains redundant components, \emph{e.g.,} task-irrelevant abilities or noisy elements \citep{skorobogat2025subspace,yadav2023ties,yu2024language}. 
As a result, there still exist conflicts in the subsequent merging procedure using these task vectors. 
DARE \citep{yu2024language} proposes to sparsify task vectors by randomly dropping elements and rescaling. Their task vector can be formulated as:
\begin{equation*}
    \Delta W_{\text{DARE}} = \frac{1}{1-p} \text{Drop}(W_{\text{FT}} - W_{\text{B}}),
\end{equation*}
where $\text{Drop}$ denotes the Dropout operation with rate $p$.
Despite its effectiveness in reducing conflict, several limitations remain as follows.
\textbf{First}, the Dropout operation involves randomness and lacks knowledge-awareness, and thus cannot locate the valuable components in a targeted manner.
\textbf{Second}, task-specific abilities actually lie in the subspace of pre-trained parameters, suggesting that binary masking in the parameter space may not truly separate redundancy or irrelevant components from the required ability.
This analysis motivates us to resort to subspace to purify task vectors. 

\begin{figure*}
\begin{center}
\includegraphics[width=0.9\linewidth]{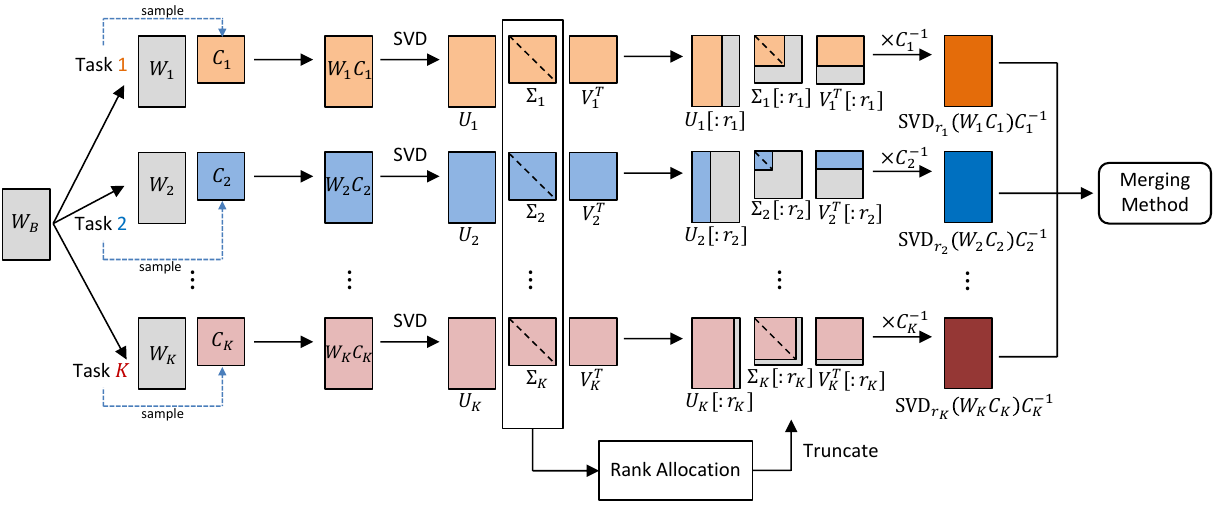} 
\end{center}
\caption{An overall illustration of our proposed method, PAVE, which serves as a plug-and-play method for task-vector based merging methods. It begins by identifying task-aware components through a decomposition approach, and then eliminates noise and less effective components via rank-based pruning. A rank allocation strategy is designed to optimize the preserved rank ratios across individual models. Overall, rather than employing the plain task vector $W_{k} - W_{\text{B}}$, PAVE 
performs model merging based on purified task vectors $\text{SVD}_{r_k}(W_{k}C_k)C^{-1}_k - W_{\text{B}},\ 1\le k \le K$.}
\end{figure*}

\subsection{Purifying Task Vectors in Knowledge-Aware Subspace}
\label{section:Purifying task vectors in knowledge-aware subspace}

Directly performing singular value decomposition (SVD) on fine-tuned model weights constructs a subspace, 
but such subspace is not necessarily aligned with task-specific knowledge, as it only reflects the spectral distribution of the parameter matrix itself
and pre-trained LLM weights handle a broad range of tasks.
To this end, we leverage the context-oriented singular value decomposition (CO-SVD) \citep{yang2024corda} to decompose fine-tuned weights into a knowledge-aware subspace. CO-SVD requires a small number of task-domain samples to construct the activation covariance matrix used for decomposition, we will further discuss this sample requirement at the end of this section.
Concretely, for each task $T_k, 1\le k\le K$, we sample some training samples from this task and feed them into the fine-tuned model $\mathcal{M}_k$. 
After acquiring the input activation of each layer $X$ (layer index is omitted for simplicity), we calculate the covariance matrix as $C=XX^T$.
Projecting the weight matrix of a linear layer onto the covariance matrix of its input activation can accentuate the task-specific components related to the input. 
Therefore, we apply SVD on the product of each weight matrix and its corresponding covariance matrix.
The decomposed components are ordered by the singular values that reflect their contributions to the task as characterized by $C$.
Such subspace enjoys a low-rank structure, which means that the knowledge-aware abilities can be condensed into the first several leading singular values and singular vectors, as will be shown in Section~\ref{subsec:task-aware-decomp}.
Hence, we filter out the task-irrelevant components by keeping only the first $r$ components, and then reconstruct the weight matrix through right multiplication with $C^{-1}$.
Consequently, the purified weight matrix can be formulated as:
\begin{equation*}
    W_{\text{FT}}^\dagger = \text{SVD}_r(W_{\text{FT}}C)C^{-1} = \sum_{i=1}^r \sigma_iu_iv^T_i C^{-1},
\end{equation*}
where $\sigma_i$ denotes the $i$-th singular value, $u_i$ and $v_i$ are the left and right singular vectors, respectively. We use the purified fine-tuned weight matrix $W_{\text{FT}}^\dagger$ to construct our task vector as:
\begin{equation}
\label{pave:task vector}
    \Delta W_{\text{PAVE}} = W_{\text{FT}}^\dagger - W_{\text{B}} = \text{SVD}_{r}(W_{\text{FT}}C)C^{-1} - W_{\text{B}}.
\end{equation}

The effectiveness of CO-SVD in identifying task-specific directions of weight matrices has been verified by \citet{yang2024corda} for model adaptation. 
In addition, $C$ is a symmetric positive semi-definite matrix, which enables efficient inverse operation with numerical stability. 
When $C$ is not invertible, we repeatedly add positive elements on its diagonal until it is invertible. 
In \cref{subsec:task-aware-decomp}, we numerically show that the knowledge-awareness in decomposition is essential in both components analysis and knowledge preserving.

To further validate the robustness of the decomposition regarding to sampling, we introduce \cref{theorem: stability analysis} to theoretically analyze the discrepancy between the sampled activation $C$ and an ideal activation $C_\star$. 
\begin{theorem}
\label{theorem: stability analysis}
Suppose $C_\star$ is an ideal symmetric positive definite activation matrix for knowledge preserving in the decomposition of $W\in\mathbb{R}^{m\times n}$. Let $W^{\dagger}_r(C) = \text{SVD}_r(WC)C^{-1}$. Assuming
\begin{equation*}
    \sigma_r(WC_\star) > \sigma_{r+1}(WC_\star).
\end{equation*}
Then there exist constants $K>0$ and $\epsilon>0$ such that for every SPD matrix $C$ satisfying
\begin{equation*}
    \|C-C_\star\|_2 \le \epsilon,
\end{equation*}
we have
\begin{equation*}
    \|W^{\dagger}_r(C)-W^{\dagger}_r(C_\star)\|_F \le K \|C-C_\star\|_2,
\end{equation*}
where $K$ is a constant independent of $\|C-C_\star\|_2$. 
\end{theorem}
\cref{theorem: stability analysis} guaranties that the task vector generated by $C$ through sampling does not deviate significantly from the ideal knowledge-preserving vector generated by $C_\star$. This observation also agrees with the numerical results in \cref{section:Analysis of sampling number and time efficiency}, where the standard deviation of the performance remains small with respect to the number of samples.
The proof is included in Appendix~\ref{section: Theoretical analysis and proofs}.

Even though the decomposition requires training samples, which may raise concerns about data accessibility, we emphasize that being data-free is not a strict requirement in the model-merging paradigm. In practice, recent works including Fisher-Merging \citep{matena2022merging} RegMean \citep{jin2022dataless}, AdaMerging \citep{yang2023adamerging} and Concrete Subspace Learning \citep{tang2023concrete} explicitly rely on training data samples or even test data for optimization. Therefore, using a limited number of unlabeled data samples is widely adopted in the field. We further discuss this problem in~\cref{subsection: data accessibility}.




\subsection{Spectral Rank Allocation}

\begin{algorithm}[t]
	\small
	\caption{
		Spectral rank allocation strategy
	}
	\label{alg1}
	\begin{algorithmic}[1]
		\REQUIRE Models $\{\mathcal{M}\{W_1^l\}_{l=1}^L,\cdots, \mathcal{M}\{W_K^l\}_{l=1}^L\}$, data sampling from target tasks $\{X_1,X_2,\cdots, X_K\}$, a preserved rank ratio $\rho$, a stopping ratio $\gamma$.
		\STATE Feed $X_i$ into $\mathcal{M}\{W_i^l\}_{l=1}^L$. Calculate covariance matrices for all layers and for all models, $\{C_i^l\}_{i, l}^{K, L}$, and the rank of each layer $\{R^l\}_{l=1}^L$.
		\FOR{$l=1$ to $L$}
            \STATE Apply SVD to $W_i^lC_i^l$ and compute $\{\sigma^l_{i,j}\}_{i,j}^{K,R^l}$, sort them in descending order. 
            \STATE Normalize $\{\sigma^l_{i,j}\}_{i,j}^{K,R^l}$ by the largest singular value, $ \{s^l_{i,j}\}_{i,j}^{K,R^l} =\left\{\frac{\sigma^l_{i,j}}{\sigma^l_{i,\max}}\right\}_{i,j}^{K,R^l}$
            \STATE Initialize $\{r_i^l \}^{K, L}_{i,l} = \{R^l \}^{K, L}_{i,l}$
            \STATE Initialize index set $I = \{1, 2, \cdots, K\}$ 
		\WHILE{$\frac{\sum_{i=1}^K r_i^l}{K R^l} > \rho$}
            \STATE Pick up the smallest $s^l_{i, r_i^l}$ across $i\in I$, suppose it is $s^l_{t, r_t^l}$
            \STATE $r_t^l = r_t^l-1$
            \IF{$r_t^l \leq \gamma R^l$}
            \STATE remove $t$ from $I$
            \ENDIF
            \ENDWHILE
		\ENDFOR
		\STATE Output: the rank ratios for each individual model, $\{\rho_i\}_{i}^K = \left\{ \frac{\sum_{l=1}^L r_i^l}{\sum_{l=1}^L R^l}\right\}_{i}^K$.
	\end{algorithmic}
\end{algorithm}

\begin{figure*}[t]
	\centering
	\begin{subfigure}{0.27\textwidth}
		\centering
		\includegraphics[width=1.\linewidth]{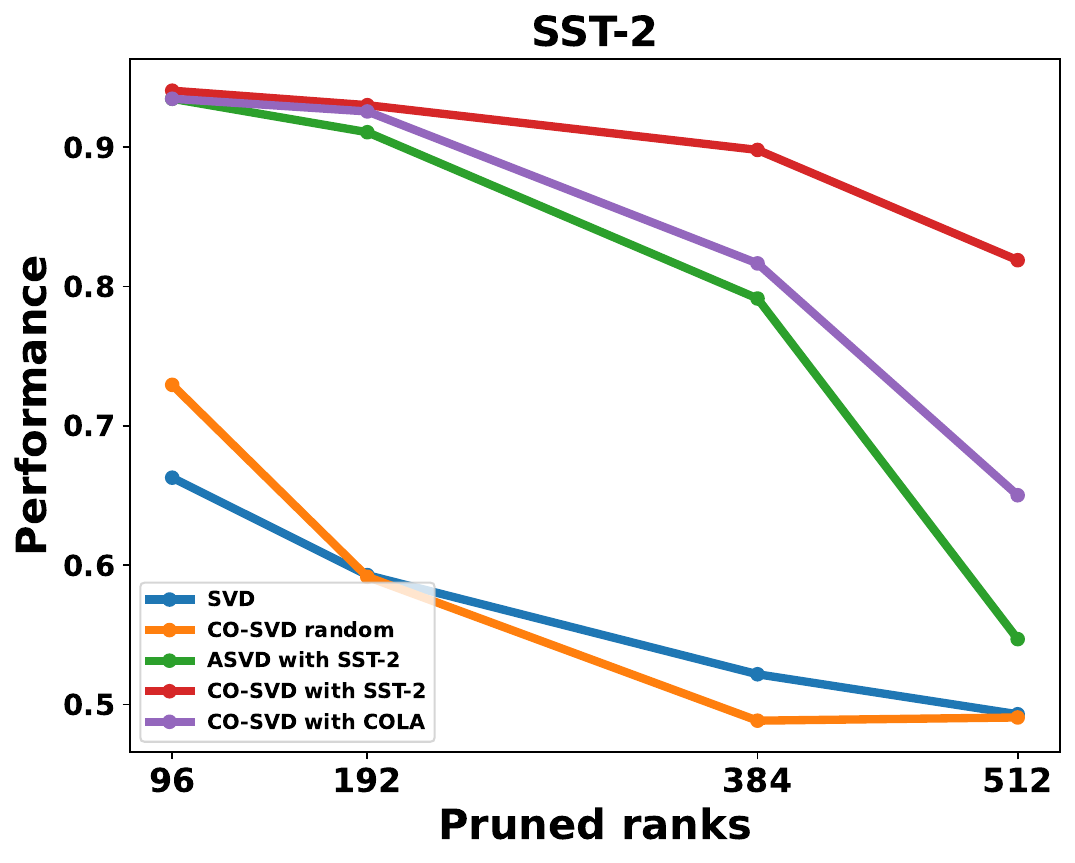}
	\end{subfigure}
	\begin{subfigure}{0.27\textwidth}
		\centering
		\includegraphics[width=1.\linewidth]{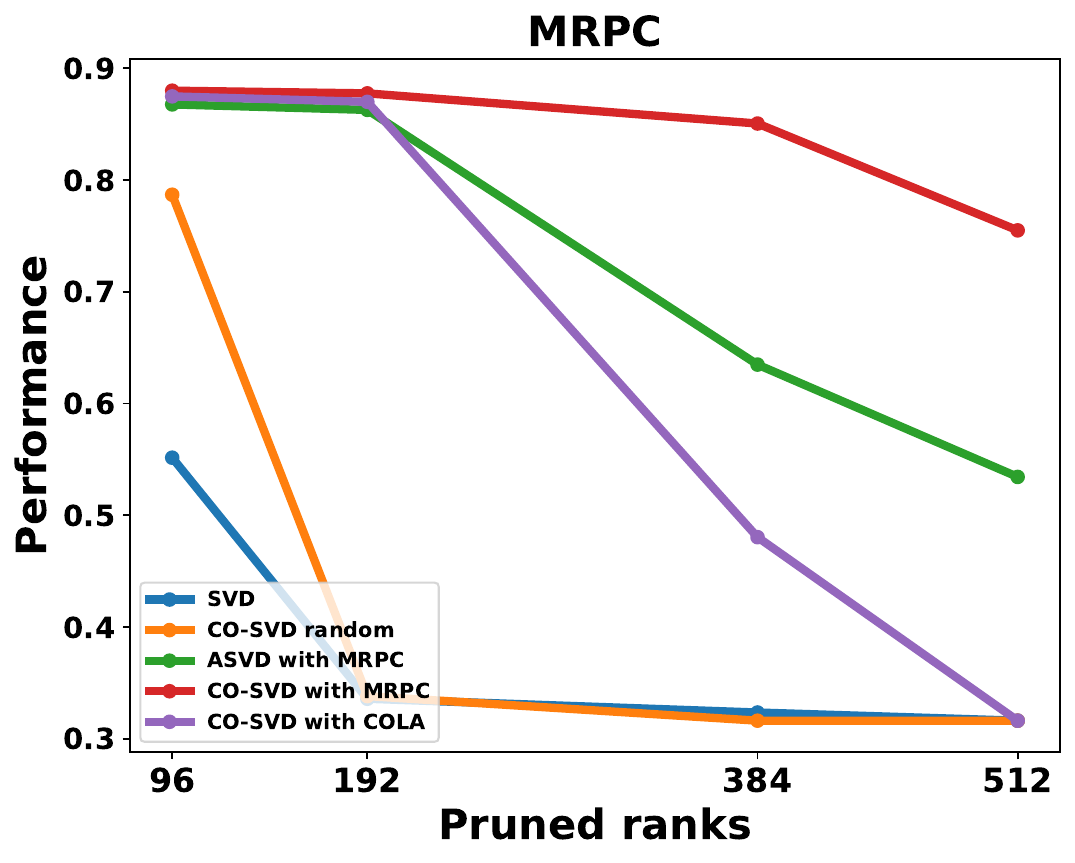}  
	\end{subfigure}
	\begin{subfigure}{0.27\textwidth}
		\centering
		\includegraphics[width=1.\linewidth]{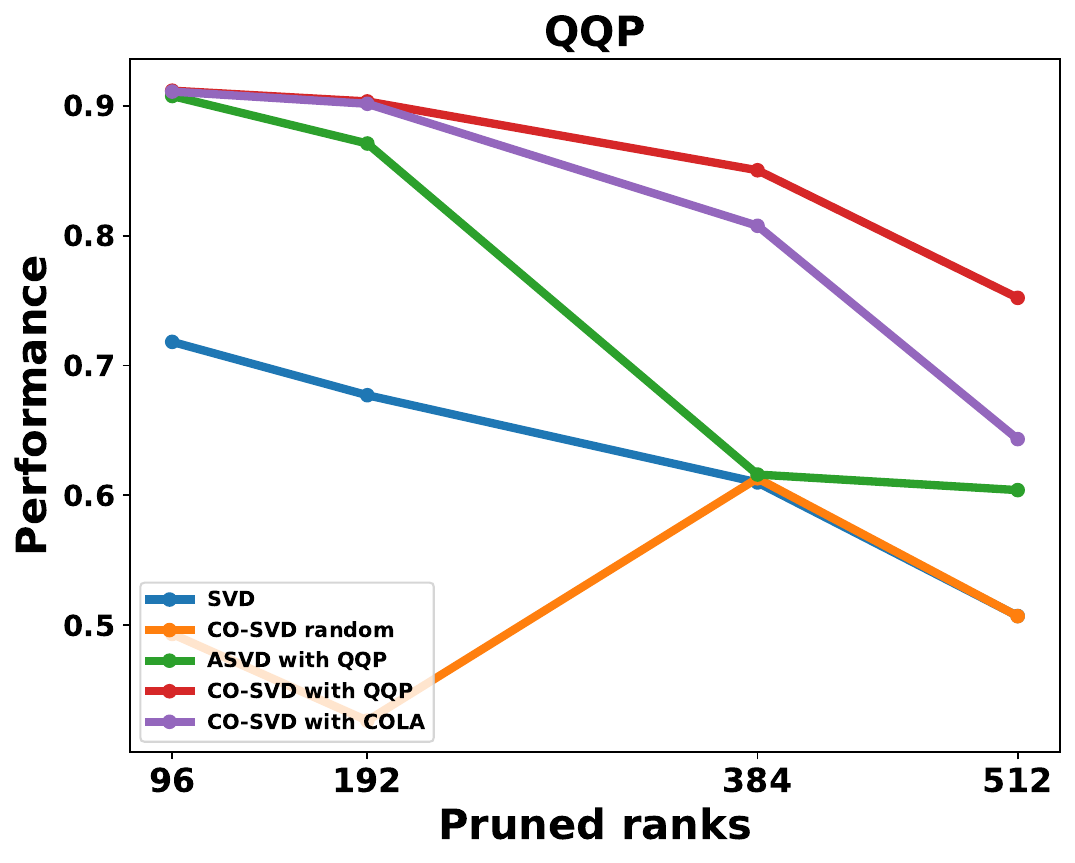}  
	\end{subfigure}	
    \begin{subfigure}{0.27\textwidth}
		\centering
		\includegraphics[width=1.\linewidth]{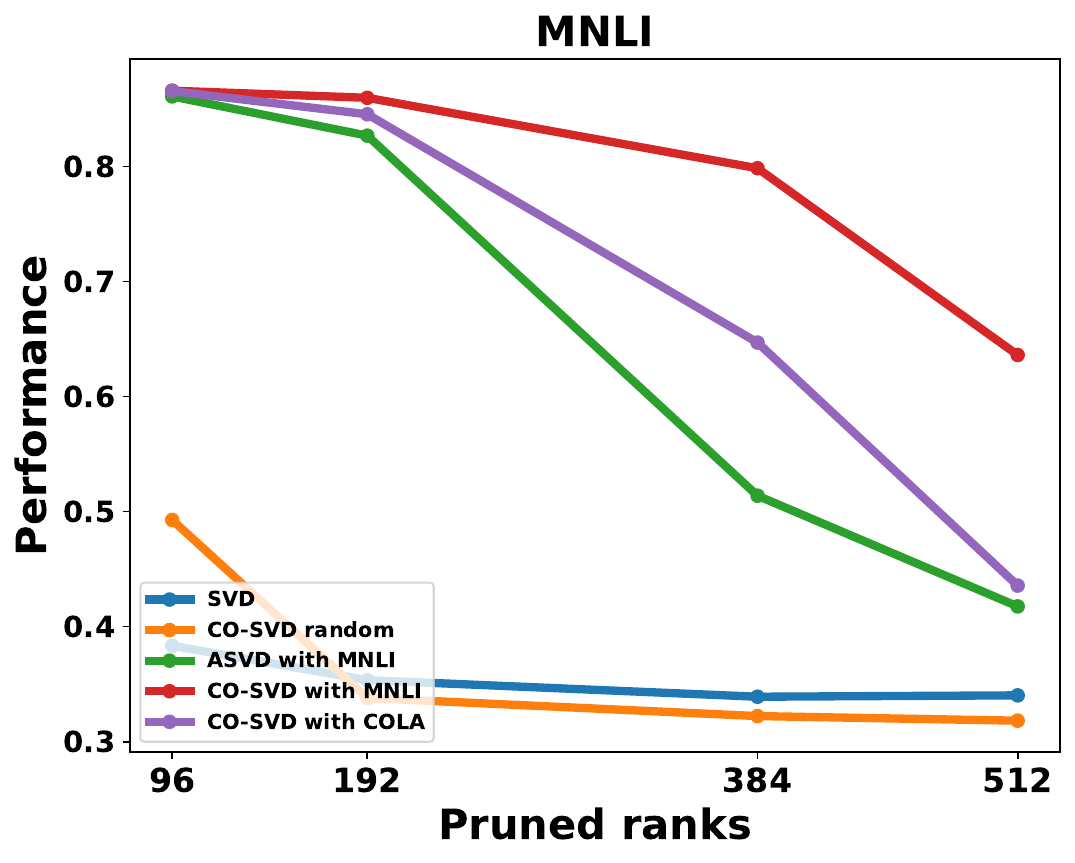}  
	\end{subfigure}	
    \begin{subfigure}{0.27\textwidth}
		\centering
		\includegraphics[width=1.\linewidth]{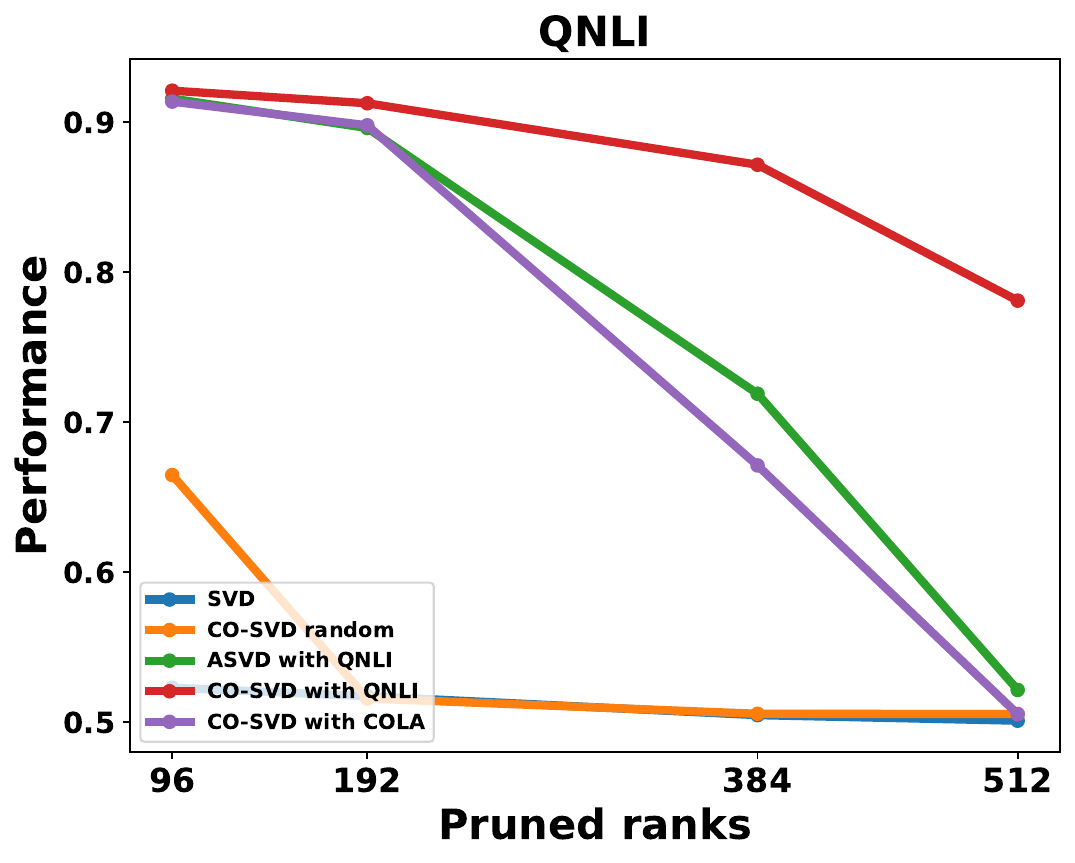}  
	\end{subfigure}
    \begin{subfigure}{0.27\textwidth}
		\centering
		\includegraphics[width=1.\linewidth]{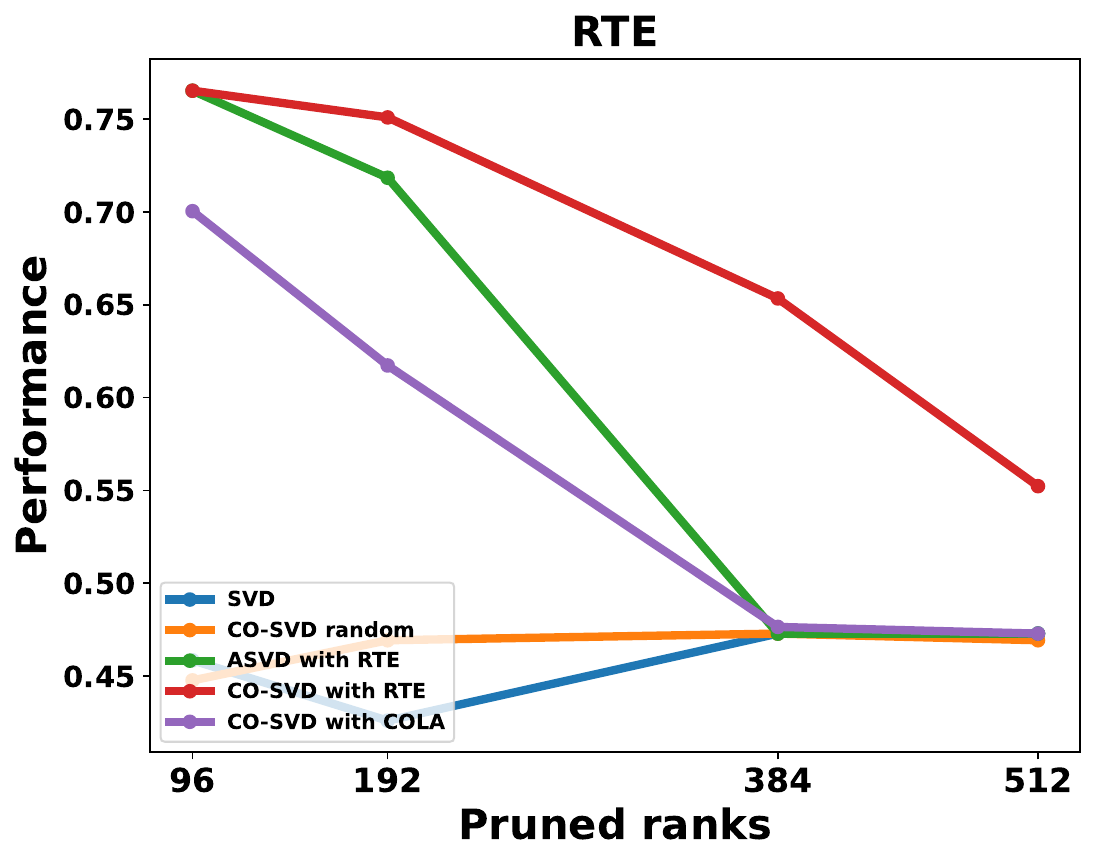}  
	\end{subfigure}
	\caption{Model rank pruning with different decomposition methods. The results show that the context-oriented decomposition (CO-SVD) is more effective in aligning the subspace with the target knowledge. The pruned ranks, 96, 192, 384 and 512, correspond to removing 1/8, 1/4, 1/2 and 2/3 of the full rank, respectively.}
    \label{graph:compression}
\end{figure*}

We have introduced the construction of our purified task vector in Section~\ref{section:Purifying task vectors in knowledge-aware subspace}. 
However, when merging multiple fine-tuned models, assigning the rank $r$ to each model is highly non-trivial. This is because the fine-tuned individual models vary in their adaptation to rank pruning; some fine-tuned weights are more robust and can preserve performance even after aggressive rank pruning, while others are more sensitive. As a result, applying a naive rank strategy can lead to performance degradation.
To optimize rank allocation under a specified preserved rank ratio $\rho$, we propose a layer-wise optimization framework that determines the preserved rank of linear layer $l$ by minimizing a summation of normalized activated pruning error from different models. 
This strategy enables an adaptive preservation of task-specific knowledge during the merging process.

We introduce the activated pruning error with rank $r^l_i$ as: 
\begin{equation}
 \|\text{SVD}_{r^l_i} (W_i^l C_i^l) - W_i^l C_i^l \|_F^2,
\end{equation}
where $i$ refers to the $i$-th model. This error estimates the activated distance between our purified task vector and the original plain task vector. Since the plain task vector retains the full capabilities through fine-tuning, this error serves as a measure of how well the purified task vector preserves task-specific knowledge. 
In practice, we further normalize it with $\sigma_{\max}(W^l_i C^l_i)$ to ensure comparability 
across different models with various spectral magnitudes.
This normalization bounds the scaled spectrum within a range 
from the inverse of its condition number to 1,
thus enabling
a fair assignment of rank across models. 
The normalized activated pruning error is defined as:
\begin{equation*}
    \frac{\|\text{SVD}_{r^l_i} (W^l_i C^l_i) - W^l_i C^l_i \|_F^2}{(\sigma_{\max}(W^l_i C^l_i))^2}  = \sum_{j=r^l_i+1}^{R^l} \left(\frac{\sigma^l_{i,j}}{\sigma^l_{i,\max}}\right)^2.
\end{equation*}
Building on this formulation, we define the overall objective for the $l$-th layer as: 
\begin{equation}
\label{loss: rank_strategy}
    \min_{\frac{\sum_{i=1}^K r^l_i}{K R^l} < \rho} \sum_{i=1}^K \frac{\|\text{SVD}_{r^l_i} (W^l_i C_i^l) - W^l_i C^l_i \|_F^2}{(\sigma_{\max}(W^l_i C^l_i))^2},
\end{equation}
where $R^l$ refers to full rank.
Minimizing the objective \cref{loss: rank_strategy} can be effectively approached using a greedy algorithm, and the whole procedure has been summarized in Algorithm~\ref{alg1}. 
In this procedure, for each layer $l$, we iteratively remove the component associated with the smallest normalized singular value, defined as $s^l_{i,j} = \frac{\sigma^l_{i,j}}{\sigma^l_{i,\max}}$, thereby achieving optimality of \cref{loss: rank_strategy} through the greedy process. 
To prevent excessive rank pruning that causes performance degradation, we introduce a threshold $\gamma$ as a stopping ratio. This rank strategy enables spectrum-based rank allocation while respecting the global rank constraint.  

\begin{table*}[t]
\renewcommand\arraystretch{1.}
\begin{center}
\small
\resizebox{\textwidth}{!}{%
\begin{tabular}{c|cccccccc|l}
\toprule[1.5pt]
{Methods} & {CoLA} & {SST-2} &  {MRPC} & {STSB} & {QQP} & {MNLI} & {QNLI} & {RTE} & {Average}\\
 \midrule
 Individual & 60.18 & 94.04 & 89.22 & 90.63 & 91.41 & 87.20 & 92.71 & 79.06 & 85.55\\
\midrule
RegMean  & 36.67 & 90.6 & 75.74 & 62.68 & 83.55 & 70.02 & 82.35 & 58.48  & 70.01 \\
DOGE-Merging & 28.64 & 88.53 & 81.13 & 78.46 & 82.34 & 59.53 & 77.27 & 67.87 & 70.46 \\
ISO-Merging  & 28.47 & 86.93 & 80.64 & 67.63 & 84.74 & 65.23 & 78.75 & 73.29  & 70.71 \\
 \midrule
EMR-Merging & 39.96  & 93.35  & 86.27 & 82.77 & 89.72 & 85.45 & 89.57 & 74.37 & 80.18\\
EMR-Merging W/ DARE & 40.58 & 93.23  & 86.52 & 82.86 & 89.93 & 85.98 & 89.68 & 75.09 & 80.48$^{\color{purple}{+0.30}}$ \\
\rowcolor{gray!15} EMR-Merging W/ PAVE & 60.42 & 93.69 & 87.75 & 89.73 & 89.13 & 85.27 & 92.09 & 76.17 &  \textbf{84.28}$^{\color{purple}{+4.10}}$ \\
 \midrule
Ties-Merging  & 20.48 & 84.40 & 81.13 & 58.19 & 85.70 & 64.65 & 74.81 & 42.96 & 64.04 \\
Ties-Merging W/ DARE & 17.56 & 83.83 & 80.39 &  63.60 & 85.79 & 64.82 & 73.24 & 44.77 & 64.25$^{\color{purple}{+0.21}}$ \\
\rowcolor{gray!15} Ties-Merging W/ PAVE  & 25.45 & 82.45  & 79.41 & 64.03 & 85.43 & 69.93 & 77.19 & 51.62 & \textbf{66.93}$^{\color{purple}{+2.89}}$\\
 \midrule
WUDI-Merging & 22.13 & 89.45 & 80.15 & 61.24 & 82.35 & 65.80 & 77.85 & 65.34 & 68.04\\
WUDI-Merging W/ DARE & 23.49 & 89.45 & 80.15 & 61.24 & 82.35 & 65.80 & 77.85 & 65.34 & 68.20$^{\color{purple}{+0.16}}$ \\
\rowcolor{gray!15} WUDI-Merging W/ PAVE  & 34.77 & 88.76 & 80.39 & 68.82 & 83.26 & 71.50 & 78.58 & 64.26 &  \textbf{71.29}$^{\color{purple}{+3.25}}$ \\
\bottomrule[1.5pt]
\end{tabular}
}
\end{center}
\caption{Results of merging fine-tuned RoBERTa models on eight datasets from GLUE benchmark.}
\label{roberta results}
\end{table*}

\begin{table*}[t]
\renewcommand\arraystretch{1.0}
\begin{center}
\small
\resizebox{\textwidth}{!}{%
\begin{tabular}{c|cccccccc|l}
\toprule[1.5pt]
{Methods} & {CoLA} & {SST-2} &  {MRPC} & {STSB} & {QQP} & {MNLI} & {QNLI} & {RTE} & {Average}\\
 \midrule
 Individual & 59.10 & 95.06 & 89.21 & 91.31 & 91.50 & 88.52 & 93.30 & 69.31 & 84.66\\
\midrule
RegMean & 22.59 & 93.00 & 72.55 & 27.99 & 81.89 & 73.77 & 73.44 & 81.80 & 65.87 \\
DOGE-Merging  & 2.80 & 75.60 & 69.11 & 53.39 & 64.05 & 44.27 & 54.02 & 51.34 & 51.82 \\
ISO-Merging  & 7.42 & 60.92 & 67.35 & 44.91 & 63.12 & 65.93 & 70.01 & 45.31 & 53.12 \\
Twin-Merging & 46.13 & 93.10 & 73.24 & 77.57 & 86.23 & 72.68 & 89.86 & 61.73	 & 75.06 \\
 \midrule
EMR-Merging & 47.05  & 94.72  & 71.32 & 87.57 & 88.36 & 87.38 & 92.29 & 64.26 & 79.11\\
EMR-Merging W/ DARE & 47.76 & 94.61  & 72.06 & 87.30 & 88.77 & 87.67 & 92.28 & 66.43 & 79.61$^{\color{purple}{+0.50}}$ \\
\rowcolor{gray!15} EMR-Merging W/ PAVE  & 61.32  & 94.50  & 84.31 & 89.54 & 90.34 & 87.50 & 93.01 & 60.65 & \textbf{82.64}$^{\color{purple}{+3.53}}$\\
 \midrule
Task Arithmetic  & 3.03 & 75.00 & 68.38 & 52.72 & 63.44 & 43.46 & 51.16 & 50.90 &  51.01\\
Task Arithmetic W/ DARE  & 2.84 & 68.81 & 68.38 & 54.24 & 64.38 & 42.23 & 55.15 & 49.46 & 50.68$^{\color{ForestGreen}{-0.33}}$ \\
\rowcolor{gray!15} Task Arithmetic W/ PAVE   & 2.18  & 75.46  & 68.38 & 55.90 & 64.00 & 43.48 & 61.25 & 51.99 & \textbf{52.83}$^{\color{purple}{+1.82}}$ \\
 \midrule
Ties-Merging & 5.36 & 66.28 & 68.63 & 19.09 & 63.68 & 40.44 & 55.39 & 48.74 & 45.95 \\
Ties-Merging W/ DARE  & 6.03 & 56.42 & 68.87 & 0.61 & 66.28 & 42.11 & 54.00 & 51.26 & 43.19$^{\color{ForestGreen}{-2.76}}$ \\
Ties-Merging W/ SubBo  & 3.08 & 66.63 & 68.63 & 25.86 & 63.51	 & 41.88 & 57.66 & 50.54 & 47.22$^{\color{purple}{+1.27}}$ \\
\rowcolor{gray!15} Ties-Merging W/ PAVE  & 1.39 & 68.35 &  68.63 & 36.76 & 63.12 & 44.01 & 59.36 & 50.90 & \textbf{49.06}$^{\color{purple}{+3.11}}$ \\
\bottomrule[1.5pt]
\end{tabular}
}
\end{center}
\caption{Results of merging fine-tuned DeBERTa models on eight datasets from GLUE benchmark.}
\label{deberta results}
\end{table*}

\section{Experiments}
\label{Experiments}

\begin{table*}[t]
\begin{center}
\small
\begin{tabular}{l|ll|ll}
\toprule
\multirow{2}{*}{Methods} &
\multicolumn{2}{c|}{Mathematical Solving} &
\multicolumn{2}{c}{Code Generating} \\
\cmidrule(lr){2-3}\cmidrule(lr){4-5}
& GSM8K & MATH & Human Eval & MBPP \\
\midrule
Individual Math & 54.9 & 10.7 & - & -  \\
Individual Code & - & - & 39.6 & 30.1  \\
 \midrule
Average Merging & 32.2 & 5.0 & 29.2 & 29.0  \\
Average Merging W/ DARE & 31.9$^{\color{ForestGreen}{-0.3}}$ & 5.0$^{\color{purple}{+0.0}}$ & 29.8$^{\color{purple}{+0.6}}$ & 28.6$^{\color{ForestGreen}{-0.4}}$  \\
\rowcolor{gray!15}Average Merging W/ PAVE & \textbf{32.4}$^{\color{purple}{+0.2}}$ & \textbf{5.0}$^{\color{purple}{+0.0}}$ & \textbf{31.1}$^{\color{purple}{+1.9}}$ & \textbf{29.0}$^{\color{purple}{+0.0}}$  \\
 \midrule
Task Arithmetic & 34.3 & 4.1 & 33.5 & 27.6  \\
Task Arithmetic W/ DARE & 32.3$^{\color{ForestGreen}{-2.0}}$ & \textbf{4.4}$^{\color{purple}{+0.3}}$ & 32.3$^{\color{ForestGreen}{-1.2}}$ & \textbf{28.6}$^{\color{purple}{+1.0}}$  \\
\rowcolor{gray!15}Task Arithmetic W/ PAVE & \textbf{34.5}$^{\color{purple}{+0.2}}$ & 4.2$^{\color{purple}{+0.1}}$ & \textbf{35.4}$^{\color{purple}{+1.9}}$ & 28.0$^{\color{purple}{+0.4}}$  \\
\bottomrule
\end{tabular}
\end{center}
\caption{Results of merging fine-tuned LLaMA-2-7B models on Math and Code tasks. Model merging on generative tasks is particularly challenging, as it requires preserving the full reasoning ability across the entire generated response rather than making an instance-level prediction, and only a limited number of existing studies have explored this setting.}
\label{NLG results}
\end{table*}

In this section, we provide the results on the GLUE benchmarks using RoBERTa-base~\citep{liu2019roberta} and DeBERTa-base~\citep{he2020deberta}, as well as the generation benchmarks with LLaMA2-7B~\citep{touvron2023llama} on mathematical solving and code generation tasks. 
We evaluate PAVE with various merging methods, including Task Arithmetic~\citep{ilharco2022editing}, Ties-Merging~\citep{yadav2023ties}, EMR-Merging~\citep{huang2024emr}, and WUDI-Merging~\citep{cheng2025whoever}. We also included RegMean~\citep{jin2022dataless} as a baseline that uses training samples during the merging process, and include DOGE-Merging with Task Arithmetic~\citep{wei2025modeling}, Subspace Boost (simplified as SubBo) \citep{skorobogat2025subspace} with Ties-Merging, Twin-Merging \citep{lu2024twin} and ISO-Merging~\citep{marczak2025no} as plain SVD-based merging baselines. 
For further detailed experimental settings and hyperparameters, please refer to Appendix~\ref{section:Implementation Details}.
\subsection{Effect of Decomposition in Knowledge-Aware Subspaces}
\label{subsec:task-aware-decomp}

We demonstrate the effectiveness of decomposition in knowledge-aware subspaces through a series of numerical experiments. As concluded in Figure~\ref{graph:compression}, we prune individual RoBERTa models fine-tuned on six GLUE tasks using various decomposition methods. The decomposition methods mentioned in Figure~\ref{graph:compression} are summarized as follows: SVD~\citep{meng2024pissa} applies singular value decomposition directly to the weights. ASVD~\citep{yuan2023asvd} introduces a diagonal scaling matrix, where each diagonal entry corresponds to the average absolute value of the associated feature. CO-SVD~\citep{yang2024corda} 
leverages covariance matrices. 
In the CO-SVD random variant, the covariance matrix is replaced by a sampled random square matrix of [-1,1], serving as a task-agnostic baseline. CO-SVD with CoLA derives the covariance matrix using samples from CoLA, highlighting the impact of selecting samples from an appropriate task.  

From Figure~\ref{graph:compression}, it is evident that, decomposition methods that incorporate task information (CO-SVD, ASVD) outperform those that do not (SVD, CO-SVD random). In particular, CO-SVD demonstrates superior performance compared to ASVD, as CO-SVD preserves context information more effectively through task-specific activations.
Notably, performance advantages of CO-SVD become even more prominent at pruning larger ranks, where other methods exhibit sharp degradation, indicating CO-SVD’s 
effectiveness in condensing knowledge-awareness. This result shows that, compared with plain SVD and other methods, CO-SVD provides more accurate component analysis and more robust pruning, which highlights the advantage of our method over existing SVD-based approaches.

It is noteworthy that the input must originate from the corresponding task to construct knowledge-aware subspace.
When using CoLA samples to generate the covariance matrix, performance drops significantly across all tasks.
These results suggest knowledge-awareness is crucial for successful rank preservation, and emphasize the importance of introducing knowledge-aware subspaces for purified task vectors.

\subsection{Model Merging for GLUE Tasks}
\label{subsection: Model merging for GLUE tasks}

To evaluate the effectiveness of PAVE, we first present results on GLUE tasks using RoBERTa as the backbone model. The numerical results are summarized in Table~\ref{roberta results}. We observe that PAVE yields a notable average performance gain of +4.10\% when applied to EMR-Merging, reducing the gap to just 1.27\% from the individual models' average performance.
For specific tasks such as CoLA, PAVE significantly improves the score from 39.96\% to 60.42\%, even surpassing the individual model's score of 60.18\%.
These improvements are not limited to EMR-Merging. 
PAVE also shows improvements with Ties-Merging and WUDI-Merging.
For example, when integrating on WUDI-Merging, PAVE achieves an average gain of +3.25\%, outperforming the competitive baselines including RegMean, DOGE-Merging, and ISO-Merging, as well as the DARE-enhanced variant.

Similar to the experimental setting for RoBERTa, we evaluate our method on DeBERTa and present the results in Table~\ref{deberta results}.
PAVE cooperates effectively with EMR-Merging, achieving a substantial average performance gain of +3.53\%, narrowing the average gap to 2.02\% compared to the 
averaged performance (84.66\%) of individual models.
Notably, on CoLA, PAVE boosts performance to a new high of 61.32\%, even surpassing the individual model’s score of 59.10\%, also achieving an improvement of 14.27\% over the EMR-Merging baseline.
Beyond EMR-Merging, PAVE also improves results in Task Arithmetic and Ties-Merging settings.
In Ties-Merging, it achieves a considerable gain of +3.11\% over the original baseline.

\subsection{Model Merging for Generative Tasks}
\label{Model Merging for generative tasks}
For generative tasks, we conduct experiments merging the math model WizardMath-7B-V1.0~\citep{luo2023wizardmath} and the code model LLaMA-2-7B-EvolCodeAlpaca~\citep{agarwalla2024enabling}. As shown in Table~\ref{NLG results}, PAVE demonstrates a marked improvement of +1.9\% on Human Eval compared to both baselines.
Notably, PAVE achieves the highest Human Eval score (35.4\%) among all merging strategies, demonstrating the effectiveness of our purified task vectors in code generation tasks.
In the mathematical solving tasks, PAVE consistently maintains or modestly improves performance across both GSM8K (+0.2\%) and MATH (+0.1\%) compared to the Task Arithmetic baseline.
It is worth noting that, in contrast to the randomness in DARE, PAVE is more stable and practical, as none of the merged models show any degradation on the individual tasks. 
This stability, combined with consistent improvements across multiple tasks, highlights the robustness and practical advantage of PAVE. 

While PAVE consistently avoids degradation, its absolute gains on generative tasks are more modest than those observed on GLUE. This may reflect the greater difficulty of merging generative capabilities rather than reduced localization of such capabilities. To disentangle these factors, we additionally evaluate rank pruning on WizardMath-7B-V1.0 in \cref{subsection: Pruning Analysis on Generative Tasks}. CO-SVD substantially outperforms plain SVD under aggressive pruning, suggesting that knowledge-aware decomposition remains effective at localizing useful components in generative models, while the remaining challenge lies primarily in combining multiple generative capabilities.


\section{Conclusion}
In this paper, we propose the purified task vector (PAVE) as a plug-in method for task vector-based merging approaches. PAVE employs context-oriented decomposition to identify and prune redundancies of the task vectors in knowledge-aware subspace, thereby mitigating conflicts arising from redundancy. To further optimize the rank allocation across different fine-tuned models and improve the overall merging performance, we introduce a spectral rank allocation strategy that optimizes the summation of normalized activated pruning errors. In experiments, we demonstrate the effectiveness of PAVE with a wide range of merging methods, tasks and model architectures. Furthermore, we explore the integration of alternative decomposition techniques within the PAVE framework and confirm its compatibility with existing model merging approaches.

\section*{Limitations}

While PAVE demonstrates consistent improvements across various merging methods, model architectures, and datasets, it still has two main limitations. First, PAVE relies on task-specific data samples. Although many existing merging methods also use data samples in their merging procedures, and we provide a detailed discussion on data accessibility in \cref{subsection: data accessibility} together with two possible solutions when task data are unavailable, PAVE is still not a strictly data-free plug-in method for model merging. This reliance weakens its appeal in strictly data-free merging scenarios.

Second, compared with methods that mainly modify weights through element-wise masking, PAVE introduces additional computational overhead due to the SVD-based decomposition. Although the overall computational cost of PAVE remains acceptable in practice, as shown in \cref{graph: std and time}, it is still more expensive than mask-based methods.

\section*{Acknowledgments}

The research reported in this publication was supported by funding from King Abdullah University of Science and Technology (KAUST) - Center of Excellence for Generative AI, under award number 5940. For computer time, this research used Ibex managed by the Supercomputing Core Laboratory at King Abdullah University of Science and Technology (KAUST) in Thuwal, Saudi Arabia.

\bibliography{custom}

\appendix

\section{Implementation Details}
\label{section:Implementation Details}

\begin{table*}[t]
\renewcommand\arraystretch{1.}
\begin{center}
\resizebox{\textwidth}{!}{%
\begin{tabular}{c|cccccccc|c}
\toprule[1.5pt]
{Methods} & {CoLA} & {SST-2} &  {MRPC} & {STSB} & {QQP} & {MNLI} & {QNLI} & {RTE} & {Average}\\
 \midrule
 Individual & 59.10 & 95.06 & 89.21 & 91.31 & 91.50 & 88.52 & 93.30 & 69.31 & 84.66\\
 \midrule
EMR-Merging & 47.05  & 94.72  & 71.32 & 87.57 & 88.36 & 87.38 & 92.29 & 64.26 & 79.11\\
EMR-Merging W/ PAVE  plain & 59.88  & 94.61  & 82.60 & 88.83 & 90.40 & 87.35 & 92.44 & 58.48 & 81.82 \\
\rowcolor{gray!15}EMR-Merging W/ PAVE  & 61.32  & 94.50  & 84.31 &89.54 &90.34 & 87.50 & 93.01 & 60.65 &\textbf{82.64}\\
 \midrule
Task Arithmetic & 3.03 & 75.00 & 68.38 & 52.72 & 63.44 & 43.46 & 51.16 & 50.90 &  51.01\\   
Task Arithmetic W/ PAVE plain & 2.56 & 74.73 & 68.38 & 54.83 & 64.03 & 42.84 & 60.19 & 51.16 & 52.34 \\
\rowcolor{gray!15}Task Arithmetic W/ PAVE   & 2.18  & 75.46  & 68.38 & 55.90 & 64.00 & 43.48 & 61.25 & 51.99 & \textbf{52.83} \\
 \midrule
Ties-Merging  & 5.36 & 66.28 & 68.63 & 19.09 & 63.68 & 40.44 & 55.39 & 48.74 & 45.95 \\   
Ties-Merging W/ PAVE  plain & 1.04 & 70.07 &  68.87 & 31.33 & 63.16 & 43.89 & 57.79 & 50.18 & 48.29 \\
\rowcolor{gray!15}Ties-Merging W/ PAVE  & 1.39 & 68.35 &  68.63 & 36.76 & 63.12 & 44.01 & 59.36 & 50.90 & \textbf{49.06} \\
\bottomrule[1.5pt]
\end{tabular}
}
\end{center}
\caption{Results of merging fine-tuned DeBERTa models on eight datasets from GLUE benchmark.}
\label{deberta rank strategy ablation}
\end{table*}

\begin{table*}[t]
\renewcommand\arraystretch{1.1}
\label{deberta decomposition ablation}
\begin{center}
\resizebox{\textwidth}{!}{%
\begin{tabular}{c|cccccccc|c}
\toprule[1.5pt]
{Methods} & {CoLA} & {SST-2} &  {MRPC} & {STSB} & {QQP} & {MNLI} & {QNLI} & {RTE} & {Average}\\
 \midrule
 Individual & 59.10 & 95.06 & 89.21 & 91.31 & 91.50 & 88.52 & 93.30 & 69.31 & 84.66\\
 \midrule
EMR-Merging & 47.05  & 94.72  & 71.32 & 87.57 & 88.36 & 87.38 & 92.29 & 64.26 & 79.11\\
EMR-Merging W/ SVD   & 62.20 & 93.58 & 83.82 & 89.40 & 89.45 & 76.72 & 92.53 & 58.48 &  80.77 \\
EMR-Merging W/ ASVD  & 62.33 & 95.07 & 84.56 & 89.73 &89.98 &87.44 & 92.86 & 61.01 & \textbf{82.87} \\
EMR-Merging W/ SVD-LLM  & 61.82 & 94.72 & 84.07 & 89.55 & 90.10 & 87.44 & 92.90 & 61.37 &  82.74 \\
EMR-Merging W/ PAVE & 61.32 & 94.50  & 84.31 & 89.54 & 90.34 & 87.50 & 93.01 & 60.65 & 82.64\\
 \midrule
Task Arithmetic & 3.03 & 75.00 & 68.38 & 52.72 & 63.44 & 43.46 & 51.16 & 50.90 &  51.01\\  
Task Arithmetic W/ SVD  & 2.96 & 70.30 & 68.38 & 37.90 & 64.30 & 42.88 & 67.01 & 53.07 & 50.85 \\
Task Arithmetic W/ ASVD  & 2.53 & 76.03 & 66.38 & 55.60 & 64.08 & 43.23 & 61.27 & 51.62 & 52.59 \\
Task Arithmetic W/ SVD-LLM  & 2.74 & 74.89 & 68.38 & 55.48 & 64.02 & 43.54 & 61.45  & 51.62 & 52.76 \\
Task Arithmetic W/ PAVE & 2.18  & 75.46  & 68.38 & 55.90 & 64.00 &43.48 & 61.25 & 51.99 & \textbf{52.83} \\
 \midrule
Ties-Merging  & 5.36 & 66.28 & 68.63 & 19.09 & 63.68 & 40.44 & 55.39 & 48.74 & 45.95 \\   
Ties-Merging W/ SVD  & -3.29 & 53.90 & 68.38 & 6.15 & 62.98 & 35.34 & 54.22 & 52.71 & 41.29 \\
Ties-Merging W/ ASVD  & -1.74 & 66.86 & 68.63 & 17.19 & 62.84 & 40.33 & 53.10 & 46.57 & 44.22 \\
Ties-Merging W/ SVD-LLM  & 1.13 & 67.89 & 68.63 &23.33 & 63.09 & 43.02 & 57.84 & 48.74 & 46.70 \\
Ties-Merging W/ PAVE& 1.39 & 68.35 &  68.63 & 36.76 & 63.12 & 44.01 & 59.36 & 50.90 & \textbf{49.06} \\
\bottomrule[1.5pt]
\end{tabular}
}
\end{center}
\caption{Results of merging fine-tuned DeBERTa models on eight datasets from GLUE benchmark.}
\end{table*}

\subsection{Model Merging on GLUE Benchmarks}
In Section~\ref{subsection: Model merging for GLUE tasks}, we employ RoBERTa-base~\citep{liu2019roberta} and DeBERTa-base~\citep{he2020deberta} as the base model.  The performance of each method is evaluated by eight tasks in the GLUE benchmark tasks, including CoLA~\citep{warstadt2019neural}, SST-2~\citep{socher2013recursive}, MRPC~\citep{dolan2005automatically}, STS-B~\citep{cer2017semeval}, QQP~\citep{iyer2017first}, MNLI~\citep{williams2017broad}, QNLI~\citep{rajpurkar2016squad}, RTE~\citep{giampiccolo2007third}. For evaluation, CoLA is assessed using the Matthews correlation coefficient, STS-B is measured by the average of Pearson and Spearman correlation coefficients, and the remaining tasks are evaluated based on accuracy. The RoBERTa checkpoints used for model merging are publicly available via the links in~\citep{huang2024emr}. The DeBERTa checkpoints are obtained by training DeBERTa-base individually on each GLUE task for 5 epochs using the AdamW~\citep{loshchilov2017decoupled} optimizer with a learning rate of $2\times10^{-5}$, a warm-up phase of 100 steps, and a weight decay of $10^{-2}$.

Following the setting in~\citet{huang2024emr}, we report the numerical experiments of PAVE with various merging methods including Task Arithmetic~\citep{ilharco2022editing}, Ties-Merging~\citep{yadav2023ties} and EMR-Merging~\citep{huang2024emr}. We also evaluate these methods with plug-in method DARE~\citep{yu2024language}. Following the standard setting of each method, the hyperparameter $\lambda$ in Task Arithmetic and Ties-Merging is selected from \{0.1, 0.3, 0.5, 0.7, 0.9\} based on the best performance. Specifically, for the RoBERTa experiments, we set $\lambda = 0.3$ for Task Arithmetic and $\lambda = 0.9$ for Ties-Merging; for BERT experiments, we set $\lambda = 0.1$ and $\lambda = 0.5$, respectively. The optimization in WUDI-Merging performed over 300 iterations with a learning rate of $2\times10^{-5}$. The dropout ratio for DARE is fixed to $0.1$ to avoid instability. The preserved rank ratio $\rho$ of PAVE is chosen from $\{\frac7{8}, \frac{15}{16}, \frac{31}{32}, \frac{63}{64}\}$, and the stopping ratio is set to $\gamma = \rho - \frac{1-\rho}{2}$. The number of examples used to generate context matrix is 4096.

To attain optimal performance on GLUE benchmarks, PAVE progressively assigns full rank to $\Delta W_{\text{PAVE}}$ according to the performance of the plain merging method. This design reflects the observation that certain tasks are more sensitive and cannot tolerate aggressive pruning. At each step, we apply full rank to the task whose performance is furthest from that of its individual model. This process continues until all task vectors are full rank. As a result, we obtain $K$ merged models and select the one with the highest average performance. Since model merging is not computationally expensive and requires no training, the computational cost of such procedure remains acceptable.

\subsection{Model Merging on Generation Benchmarks}
Following the setup of DARE presented in~\citet{yu2024language}, we evaluate PAVE on mathematical solving and code generation tasks, as described in Section~\ref{Model Merging for generative tasks}.
For mathematical solving, we consider GSM8K~\citep{cobbe2021gsm8k} and MATH~\citep{hendrycksmath2021}, while for code generation, we use HumanEval~\citep{chen2021evaluating} and MBPP~\citep{austin2021program}. We adopt the LLaMA-2-7B~\citep{touvron2023llama} model as the base model. For mathematical solving, we utilize the fine-tuned model WizardMath-7B-V1.0~\citep{luo2023wizardmath}, while for code generation, we employ LLaMA-2-7B-EvolCodeAlpaca~\citep{agarwalla2024enabling}. Additionally, code samples are drawn from the CodeFeedBack dataset~\citep{zheng2024opencodeinterpreter} to support covariance matrix generation. The Task Arithmetic hyperparameter $\lambda$ is selected from the set $\{0.5, 1\}$, and the dropout ratio for DARE is fixed at 0.2. The pruning rank for PAVE is chosen from $\{4, 6, 8, 16\}$. The number of examples used to generate context matrix is 512.

\section{Theoretical Analysis and Proofs}

\begin{figure}[t]
	\centering
        \newlength{\figheight}
        \setlength{\figheight}{6cm}
    
	\begin{subfigure}{0.49\textwidth}
		\centering
		\includegraphics[width=1.\linewidth]{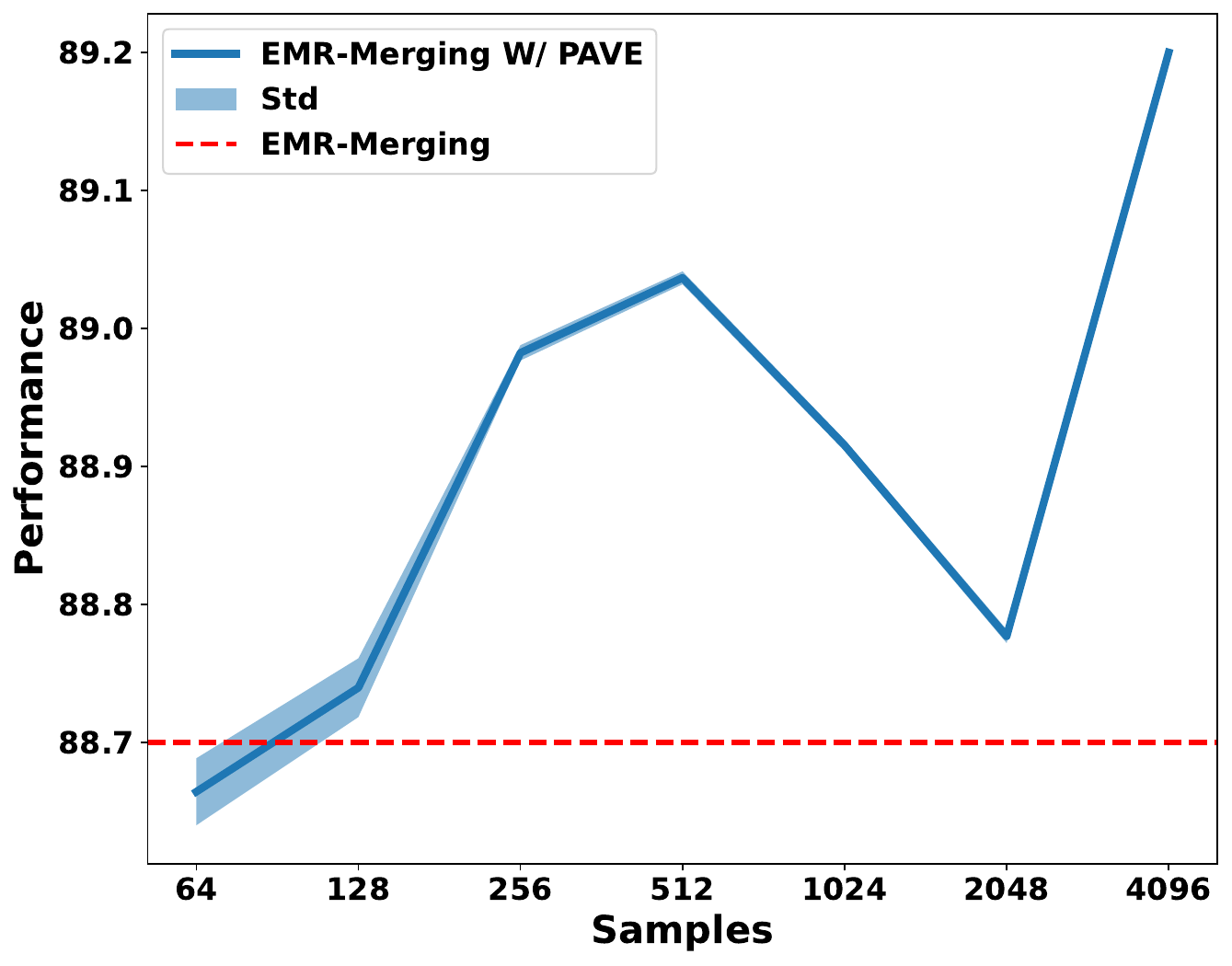}
        \label{fig:sample variance}
	\end{subfigure}
	\begin{subfigure}{0.49\textwidth}
		\centering
		\includegraphics[width=1.\linewidth]{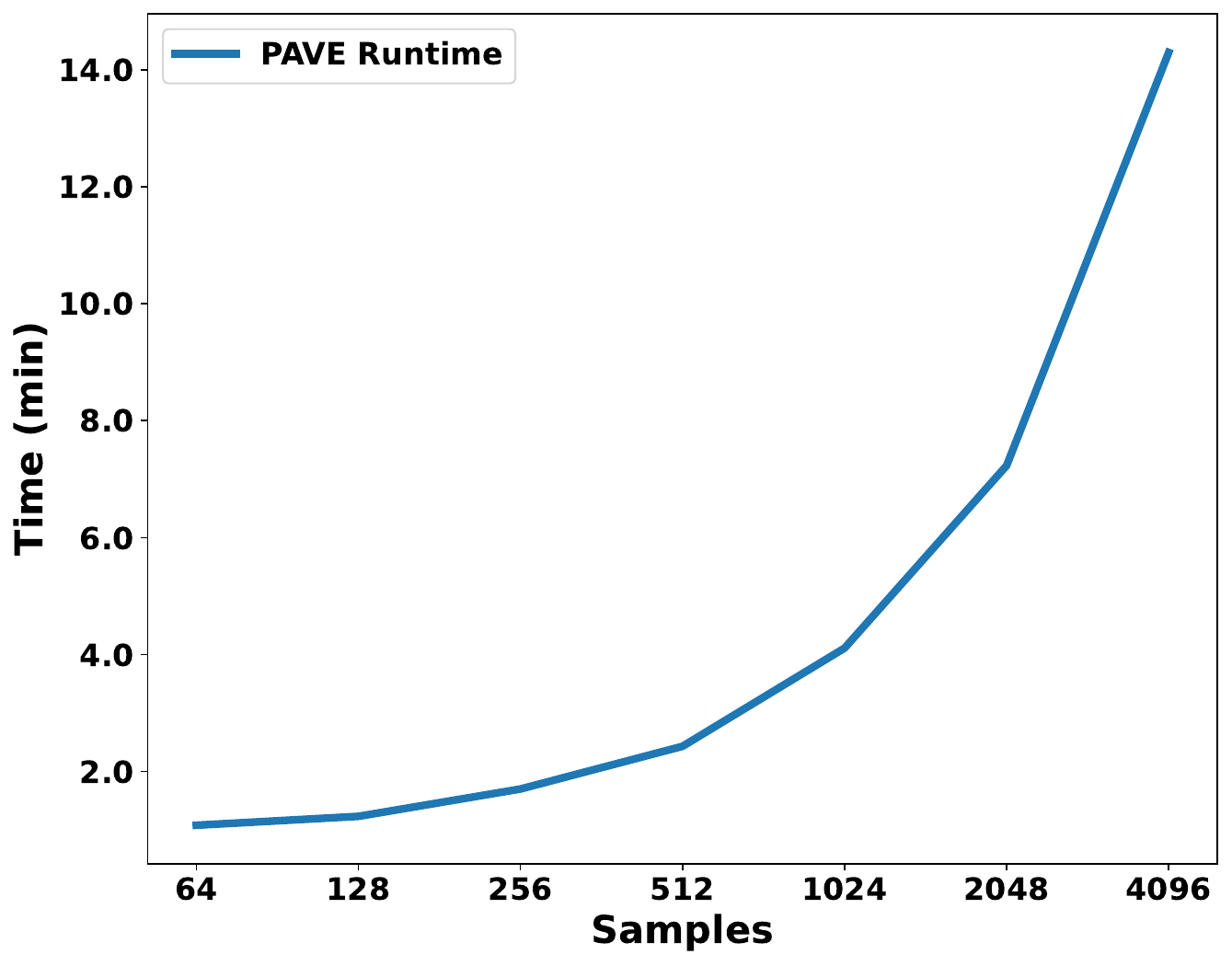}  
         \label{figure:sample time}
	\end{subfigure}
	\caption{Performance and runtime analysis of PAVE across different sample sizes. \textbf{Top:} Mean performance (solid line) and standard deviation (shaded region) over 3 random seeds of EMR-Merging with PAVE, showing consistent improvements as more samples are used. A detailed ablation has been included in \cref{vit results ablation}. \textbf{Bottom: } PAVE runtime grows with number of samples. }
    \label{graph: std and time}
\end{figure}

\begin{table*}[t]
\renewcommand\arraystretch{1.1}
\begin{center}
\resizebox{\textwidth}{!}{%
\begin{tabular}{c|cccccccc|c}
\toprule[1.5pt]
{Methods} & {SUN397} & {Cars} &  {RESISC45} & {EuroSAT} & {SVHN} & {GTSRB} & {MNIST} & {DTD} & {Average}\\
 \midrule
EMR-Merging  W/ PAVE N=4096 & 76.3$\pm<0.1$ & 74.3$\pm<0.1$ & 93.8$\pm<0.1$ & 99.3$\pm<0.1$ & 96.7$\pm<0.1$ & 97.6$\pm<0.1$ & 99.6$\pm<0.1$ & 76.1$\pm<0.1$ & 89.2$\pm<0.1$ \\
EMR-Merging  W/ PAVE N=2048 & 76.3$\pm<0.1$ & 71.5$\pm<0.1$ & 94.2$\pm<0.1$ & 99.2$\pm<0.1$ & 96.7$\pm<0.1$ & 97.6$\pm<0.1$ & 99.6$\pm<0.1$ & 75.0$\pm<0.1$ & 88.8$\pm<0.1$\\
EMR-Merging  W/ PAVE N=1024 & 76.5$\pm<0.1$ & 72.5$\pm<0.1$ & 94.2$\pm<0.1$ & 99.1$\pm<0.1$ & 96.7$\pm<0.1$ & 97.7$\pm<0.1$ & 99.6$\pm<0.1$ & 74.8$\pm<0.1$ & 88.9$\pm<0.1$\\
EMR-Merging  W/ PAVE N=512 & 76.7$\pm<0.1$ & 74.7$\pm<0.1$ & 94.0$\pm<0.1$ & 99.3$\pm<0.1$ & 96.7$\pm<0.1$ & 97.8$\pm<0.1$ & 99.5$\pm<0.1$ & 74.4$\pm<0.1$ & 89.1$\pm<0.1$ \\
EMR-Merging  W/ PAVE N=256 & 76.2$\pm<0.1$ & 74.1$\pm<0.1$ & 93.5$\pm<0.1$ & 99.1$\pm<0.1$ & 96.3$\pm<0.1$ & 97.1$\pm<0.1$ & 99.2$\pm<0.1$ & 76.3$\pm<0.1$ & 89.0$\pm<0.1$ \\
EMR-Merging  W/ PAVE N=128 & 75.8$\pm<0.1$ & 74.1$\pm<0.1$ & 93.4$\pm=0.1$ & 98.9$\pm<0.1$ & 96.4$\pm<0.1$ & 97.0$\pm<0.1$ & 99.5$\pm<0.1$ & 74.6$\pm=0.1$ & 88.7$\pm=0.1$ \\
EMR-Merging  W/ PAVE N=64 & 75.6$\pm=0.2$ & 74.3$\pm=0.1$ & 93.6$\pm=0.1$ & 98.7$\pm<0.1$ & 96.2$\pm<0.1$ & 96.5$\pm<0.1$ & 99.5$\pm<0.1$ & 74.6$\pm=0.2$ & 88.6$\pm=0.2$\\
\bottomrule[1.5pt]
\end{tabular}
}
\end{center}
\caption{Detailed ablation on \cref{graph: std and time}.}
\label{vit results ablation}
\end{table*}

\label{section: Theoretical analysis and proofs}
In this section, we present a theoretical analysis along with proofs for the theorems introduced in the previous sections.

\textbf{\cref{theorem: stability analysis}}: 
Suppose $C_\star$ is an ideal symmetric positive definite activation for knowledge preserving in the decomposition of $W\in\mathbb{R}^{m\times n}$. Let $W^{\dagger}_r(C) = \text{SVD}_r(WC)C^{-1}$. Assuming
\begin{equation*}
    \sigma_r(WC_\star) > \sigma_{r+1}(WC_\star).
\end{equation*}
Then there exist constants $K>0$ and $\epsilon>0$ such that for every SPD matrix $C$ satisfying
\begin{equation*}
    \|C-C_\star\|_2 \le \epsilon,
\end{equation*}
we have
\begin{equation*}
    \|W^{\dagger}_r(C)-W^{\dagger}_r(C_\star)\|_F \le K \|C-C_\star\|_2,
\end{equation*}
where $K$ is a constant independent of $\|C-C_\star\|_2$.

\begin{proof}
The key is to prove that $ \|W^{\dagger}_r(C)-W^{\dagger}_r(C_\star)\|_F$ is locally Lipschitz smooth at $C_\star$.
Since we have
\begin{align*}
    & W^{\dagger}_r(C)-W^{\dagger}_r(C_\star) \\
    = &  (\text{SVD}_r(WC) - \text{SVD}_r(WC_\star))C^{-1} \\
    + &  \text{SVD}_r(WC_\star) (C^{-1} - C_\star^{-1}).  
\end{align*}
Therefore, 
\begin{align*}
    & \|W^{\dagger}_r(C)-W^{\dagger}_r(C_\star)\|_F \\
    \leq &  \|\text{SVD}_r(WC) - \text{SVD}_r(WC_\star)\|_F \|C^{-1}\|_2 \\
    + &  \|\text{SVD}_r(WC_\star)\|_F \|C^{-1} - C_\star^{-1}\|_2.  
\end{align*}

Because $C_\star$ is SPD, the matrix inverse is locally Lipschitz in a neighborhood of $C_\star$. Therefore, $\|C^{-1}\|_2$ is bounded and $\|C^{-1} - C_\star^{-1}\|_2$ is locally Lipschitz at $C_\star$.

Moreover, the assumption
\begin{equation*}
    \sigma_r(WC_\star)>\sigma_{r+1}(WC_\star)
\end{equation*}
implies that the rank-$r$ truncated SVD map is locally Lipschitz at $WC_\star$. Since $C\mapsto WC$ is linear, then we have $\|\text{SVD}_r(WC) - \text{SVD}_r(WC_\star)\|_F$ is locally Lipschitz at $C_\star$. Hence $\|W^{\dagger}_r(C)-W^{\dagger}_r(C_\star)\|_F$ is locally Lipschitz at $C_\star$, which completes the proof.

\end{proof}

\section{Additional Analysis and Results}
\label{ablation study}

\subsection{Ablation Studies on Rank Allocation Strategy}

We evaluate our rank allocation strategy Algorithm~\ref{alg1} in Table~\ref{deberta rank strategy ablation}, using the same DeBERTa experiment settings on models and merging methods in Section~\ref{subsection: Model merging for GLUE tasks}. The plain method uses a fixed preserved rank ratio across all layers and models.
Compared to this baseline, our adaptive strategy consistently improves performance across different merging settings.
In EMR-Merging, the rank strategy improves the average performance from 81.82\% to 82.64\% (+0.82\%).
In other merging methods like Task Arithmetic and Ties-Merging, the gains are +0.49\% (from 52.34\% to 52.83\%) and +0.77\% (from 48.29\% to 49.06\%), respectively.
These results suggest that uniform pruning represents a naive and suboptimal approach, while our rank allocation strategy enables better performance under the same pruning budget. Additionally, we include further experiments, presenting numerical results when PAVE is integrated with alternative decomposition methods, as well as an analysis of sample size and runtime performance. Please refer to Appendix~\ref{ablation study} for the results.

\subsection{Task Vector with Various Decomposition Methods}
In this section, we present numerical experiments evaluating the performance of various decomposition methods applied to our decomposition-based task vector. The setting of models and merging methods follow the DeBERTa settings in Section~\ref{subsection: Model merging for GLUE tasks}. The evaluated decomposition methods include: The basic SVD method~\citep{meng2024pissa}, directly approximates the fine-tuned weights via a plain SVD decomposition. ASVD~\citep{yuan2023asvd}, incorporates an input-dependent diagonal scaling matrix derived from the average absolute values of the previous layer’s outputs. SVD-LLM~\citep{wang2024svd}, applies Cholesky decomposition to the previous layer’s output. CO-SVD~\citep{yang2024corda}, as the default decomposition method embedded in PAVE, utilizes the covariance matrix of the output of the previous layer as the activation matrix.
Among the evaluated methods, CO-SVD demonstrates the most consistent performance on Task Arithmetic and Ties-Merging. When the decomposition method in PAVE is switched from CO-SVD to ASVD, the resulting variant achieves the best performance with EMR-Merging. These results highlight the strong potential of our decomposition-based purified task vector framework, which is designed to integrate seamlessly with a variety of decomposition methods.

\subsection{Analysis of Sample Size and Time Efficiency}
\label{section:Analysis of sampling number and time efficiency}

\begin{table}[!t]
\begin{center}
\small
\begin{tabular}{c|l|l}
\toprule
\multirow{1}{*}{Methods} &
\multicolumn{1}{|c|}{Average Performance} &
\multicolumn{1}{c}{Time (min)} 
\\
\midrule
WUDI &  68.04 & 2.1 \\
WUDI W/ PAVE  &  71.29$^{\color{purple}{+3.25}}$   & 4.4$^{\color{purple}{+2.3}}$  \\
\bottomrule
\end{tabular}
\end{center}
\caption{Comparison of computational cost and average performance. Adding PAVE to WUDI improves the average performance while introducing an acceptable increase in computation time.}
\label{table: computational cost}
\end{table}


We next analyze the impact of sample size on the stability, efficiency, and reproducibility of PAVE. As illustrated in Figure~\ref{graph: std and time}, larger sample sizes generally reduce the variance while maintaining strong overall performance of EMR-Merging with PAVE, highlighting its stability and robustness. When the sample size is very small, due to the regularization, the method degenerates to plain SVD, which explains the lower performance observed in this regime. Importantly, we observe that even a moderate sample size (256–512) is sufficient to yield substantial performance gains, offering a practical trade-off between accuracy and computational cost. This finding suggests that PAVE can be effectively deployed in practice without incurring excessive runtime overhead, while still preserving most of the performance benefits of using larger sample sizes. Throughout our numerical experiments, we increased the number of samples up to 4096 to ensure a stable and consistent demonstration of our method’s performance. Employing larger sample sizes also helps ensure the reproducibility of our results. All runtime measurements were conducted on a workstation equipped with a single NVIDIA A100 GPU and 8 CPUs, with 512 GB of RAM.


\subsection{Data Accessibility}
\label{subsection: data accessibility}

\begin{table*}[t]
\begin{center}
\small
\begin{tabular}{l|ll|ll}
\toprule
\multirow{2}{*}{Methods} &
\multicolumn{2}{c|}{Mathematical Solving} &
\multicolumn{2}{c}{Code Generating} \\
\cmidrule(lr){2-3}\cmidrule(lr){4-5}
& GSM8K & MATH & Human Eval & MBPP \\
\midrule
Individual Math & 54.9 & 10.7 & - & -  \\
Individual Code & - & - & 39.6 & 30.1  \\
 \midrule
Average Merging & 32.2 & 5.0 & 29.2 & 29.0  \\
Average Merging W/ DARE & 31.9$^{\color{ForestGreen}{-0.3}}$ & 5.0$^{\color{purple}{+0.0}}$ & 29.8$^{\color{purple}{+0.6}}$ & 28.6$^{\color{ForestGreen}{-0.4}}$  \\
\rowcolor{gray!15}Average Merging w/ PAVE & 32.4$^{\color{purple}{+0.2}}$ & 5.0$^{\color{purple}{+0.0}}$ & 31.1$^{\color{purple}{+1.9}}$ & 29.0$^{\color{purple}{+0.0}}$  \\
\rowcolor{gray!15}Average Merging W/ PAVE related domain data & 32.3$^{\color{purple}{+0.1}}$ & 5.1$^{\color{purple}{+0.1}}$ & 31.1$^{\color{purple}{+1.9}}$ & 29.4$^{\color{purple}{+0.4}}$  \\
\rowcolor{gray!15}Average Merging W/ PAVE synthetic data & 32.2$^{\color{purple}{+0.0}}$ & 5.1$^{\color{purple}{+0.1}}$ & 30.9$^{\color{purple}{+1.7}}$ & 29.3$^{\color{purple}{+0.3}}$  \\
\bottomrule
\end{tabular}
\end{center}
\caption{Results of merging fine-tuned LLaMA-2-7B models on Math and Code tasks.}
\label{data free results}
\end{table*}

\begin{table*}[t]
\renewcommand\arraystretch{1.1}
\begin{center}
\resizebox{\textwidth}{!}{%
\begin{tabular}{c|cccccccc|l}
\toprule[1.5pt]
{Methods} & {SUN397} & {Cars} &  {RESISC45} & {EuroSAT} & {SVHN} & {GTSRB} & {MNIST} & {DTD} & {Average}\\
 \midrule
 Individual ViT-B/32 & 75.3 & 77.7 & 96.1 & 99.7 & 97.5 & 98.7 & 99.7 & 79.4 & 90.5\\
EMR-Merging& 75.2  & 72.8  & 93.5 & 99.5 & 96.9 & 98.1 & 99.6 & 74.4 & 88.7\\
\rowcolor{gray!15}EMR-Merging  W/ PAVE  & 76.3  & 74.3  & 93.8 & 99.3 & 96.7 & 97.6 & 99.6 & 76.1 & \textbf{89.2}$^{\color{purple}{+0.5}}$\\
 \midrule
 Individual ViT-B/16 & 79.0  & 87.0  & 96.4 & 99.1 & 97.7 & 99.0 & 99.7 & 82.3 & 92.5 \\
EMR-Merging& 78.5  & 82.6  & 95.4 & 99.1 & 97.6 & 98.8 & 99.6 & 78.3 & 91.2 \\
\rowcolor{gray!15}EMR-Merging  W/ PAVE  & 80.2  &  82.8 & 96.1 & 99.2 & 97.6 & 98.7 & 99.6 & 78.9 & \textbf{91.7}$^{\color{purple}{+0.5}}$\\
 \midrule
 Individual ViT-L/14 & 82.3 & 92.4 & 97.4 & 100 & 98.1 & 99.2 & 99.7 & 84.1 & 94.2\\
 AWD-Merging & 76.2 & 85.4 & 88.7 & 96.1 & 92.4 & 92.3 & 99.3 & 69.4 & 87.5 \\
 DOGE-Merging  & 70.5 & 74.8 & 88.7 & 94.1 & 91.6 & 95.7 & 98.8 & 72.5 & 85.9 \\
EMR-Merging& 83.1  & 90.7  & 96.7 & 99.7 & 97.9 & 99.1 & 99.6 & 82.7 & 93.7 \\
\rowcolor{gray!15}EMR-Merging  W/ PAVE  & 83.6  & 91.7  & 97.2 & 99.8 & 99.7 & 98.9 & 99.7 & 82.2& \textbf{94.1}$^{\color{purple}{+0.4}}$\\
\bottomrule[1.5pt]
\end{tabular}
}
\end{center}
\caption{Results of merging fine-tuned ViT models on 8 image classification benchmarks.}
\label{vit results}
\end{table*}

\begin{table*}[t]
\renewcommand\arraystretch{1.1}
\begin{center}
\resizebox{\textwidth}{!}{%
\begin{tabular}{c|cccccccccccc}
\toprule[1.5pt]
\diagbox{DeBERTa}{Layers} & 1 & 2 & 3 & 4 & 5 & 6 & 7 & 8 & 9 & 10 & 11 & 12 \\
 \midrule
CoLA & 1.03 & 1.11 & 0.97 & 1.03 & 1.01 & 1.05 & 0.94  & 0.92 & 0.90 & 0.89 & 0.91 & 0.69\\
 \midrule
 MRPC & 0.81 & 0.50  & 0.48  & 0.68 & 0.79 & 0.76 & 0.94 & 0.81 & 1.22 & 0.96 & 0.97 & 1.12 \\
 \midrule
 QQP & 1.17 & 1.17 & 1.18 & 1.13 & 1.13 & 1.14 & 1.12 & 1.17 & 1.09 & 1.14 & 1.15 & 1.01 \\
\bottomrule[1.5pt]
\end{tabular}
}
\end{center}
\caption{Normalized assigned pruning ranks across layers}
\label{table: Normalized assigned pruning ranks across layers}
\end{table*}

\begin{table*}[t]
\renewcommand\arraystretch{1.1}
\begin{center}
\resizebox{\textwidth}{!}{%
\begin{tabular}{c|cccccccc}
\toprule[1.5pt]
Task &  {CoLA} & {SST-2} &  {MRPC} & {STSB} & {QQP} & {MNLI} & {QNLI} & {RTE} \\
 \midrule
Assigned Rank / Uniform Rank & 0.96 & 1.21 & 0.84 & 1.17 & 1.13 & 1.09 & 0.87 & 0.72  \\
\bottomrule[1.5pt]
\end{tabular}
}
\end{center}
\caption{Average normalized assigned pruning ranks for different downstream tasks.}
\label{table: Average normalized assigned pruning ranks for different downstream tasks.}
\end{table*}

In this section we discuss the condition when training data is strictly unavailable. In fact, the use of a limited number of data samples is not only widely accepted but also integral to several established merging techniques. For example, Fisher-Merging \citep{matena2022merging} RegMean \citep{jin2022dataless} and RegMean++ \citep{nguyen2025regmean++} explicitly rely on training data samples, while test-time adaptation methods like AdaMerging \citep{yang2023adamerging} and Concrete Subspace Learning \citep{tang2023concrete} also require test data to optimize an entropy-based objective. 

Our method PAVE similarly relies on a small number of task-domain samples to compute the covariance matrix. Importantly, our approach does not require full access to the training set. As demonstrated in Figure~\ref{graph: std and time}, as few as 256 samples are sufficient to achieve strong performance and induce minor variance. To further address the concerns about strictly data-free settings, we additionally provide two alternative solutions that do not rely on the original training data.

\textbf{Using data from a related domain.} The samples we use serve only to activate the model’s weights during decomposition. If the original training data is no longer accessible, it is sufficient to use data drawn from a similar knowledge domain. Such insight has been shared in the prior study \citep{yang2024corda}. Following this principle, we can also use data from a similar domain without access to the source training data. Concretely, in our generative task experiments under the previous setting, the covariance matrix is generated by samples from the training set of GSM8K and MATH and CodeFeedBack. In the following experiments, we show that Lila-Math \citep{mishra2022lila} and OpenCodeInstruct \citep{ahmad2025opencodeinstruct} can effectively replace samples from the original training sets. 

\textbf{Using LLM-generated synthetic data.} When even related-domain datasets are unavailable, one can use synthetic data generated by large language models in the same knowledge domain. In practice, the prompt for synthetic data generation can be composed of a basic task description, a knowledge domain description, and a few examples. The basic task description helps the language model recognize that the prompt concerns a sample generation task, while the knowledge domain description provides information of the knowledge domain, and a few examples can improve the consistency of generation. Such synthetic data are sufficient to activate the relevant parameter subspaces for our decomposition procedure.

We include numerical results in Table~\ref{data free results}, which validate the effectiveness of the two proposed data-free solutions. Therefore, when the source training data is strictly inaccessible, one can easily use dataset in similar knowledge domain or use LLMs to generate synthetic data in similar knowledge domain, and still achieve good performance using PAVE.

\subsection{Pruning Analysis on Generative Tasks}
\label{subsection: Pruning Analysis on Generative Tasks}

To further examine whether covariance-based subspace purification is effective for generative models, we conducted an additional rank-pruning experiment on WizardMath-7B-V1.0 using GSM8K in \cref{table: generative task pruning}. The results show that CO-SVD preserves math reasoning ability better than plain SVD, especially under stronger rank pruning. This suggests that covariance-based decomposition can still localize useful task-relevant components in larger generative models, and the final merging gains remain limited by the difficulty of combining different generative abilities.

\begin{table}[!t]
\begin{center}
\small
\begin{tabular}{c|c|c}
\toprule
\multirow{1}{*}{WizardMath-7B-V1.0 on GSM8K} &
\multicolumn{1}{|c|}{SVD} &
\multicolumn{1}{c}{CO-SVD} 
\\
\midrule
Pruning rank = 16 & 48.59  & 51.32 \\
Pruning rank = 64  &  47.08   &  50.94  \\
Pruning rank = 128 & 44.50 &  50.49 \\
Pruning rank = 256 & 36.16 &  49.35 \\
\bottomrule
\end{tabular}
\end{center}
\caption{Comparison of SVD and CO-SVD under different pruning ranks for WizardMath-7B-V1.0 on GSM8K.}
\label{table: generative task pruning}
\end{table}

\subsection{Model Merging for ViT Models}
\textbf{Settings.} Following the setup of EMR-Merging presented in~\citet{huang2024emr}, we employ ViT-B/32, ViT-B/16 and ViT-L/14~\citep{radford2021learning} as the base model to evaluate the performance of PAVE across 8 image classification tasks: SUN397~\citep{xiao2010sun}, Cars~\citep{krause20133d}, RESISC45~\citep{cheng2017remote}, EuroSAT~\citep{helber2019eurosat}, SVHN~\citep{netzer2011reading}, GTSRB~\citep{stallkamp2011german}, MNIST~\citep{lecun1998mnist}, and DTD~\citep{cimpoi2014describing}. The preserved rank ratio $\rho$ of PAVE is chosen from $\{\frac7{8}, \frac{15}{16}, \frac{31}{32}, \frac{63}{64}\}$, and the stopping ratio is set to $\gamma = \rho - \frac{1-\rho}{2}$. The checkpoints are publicly accessible via the link provided in~\citet{ilharco2022editing}.

\textbf{Results.} As shown in Table~\ref{vit results}, while the plain EMR-Merging method already demonstrates strong performance against other baselines including AWD-Merging and DOGE-Merging, PAVE still consistently enhances the performance across all ViT variants.
For ViT-B/32, PAVE improves the average accuracy by +0.5\% (from 88.7\% to 89.2).
On the more capable ViT-B/16 model, PAVE again contributes an improvement of +0.5\%, increasing the average from 91.2\% to 91.7\%. 

\subsection{Layer-wise Analysis of the Spectral Rank Allocation Strategy}

To better understand the behavior of the proposed spectral rank allocation strategy, we analyze the assigned pruning ranks across different layers and tasks. We report the assigned rank normalized by the corresponding uniform-rank baseline, where a value greater than 1 indicates that more ranks are preserved than under uniform allocation, and a value smaller than 1 indicates more aggressive pruning.

\textbf{Layer-wise analysis.} We observe task-specific layer-wise patterns for some fine-tuned models. As shown in Table~\ref{table: Normalized assigned pruning ranks across layers}, the CoLA model tends to preserve relatively more ranks in the early layers, while the MRPC model exhibits the opposite behavior, allocating more ranks to the later layers. In contrast, the QQP model maintains relatively uniform rank assignments across layers. These observations suggest that different downstream tasks rely on different parts of the network.

However, these patterns are not consistent across all tasks. Averaging the assigned ranks over all fine-tuned models, we do not observe any particular layer that consistently receives higher or lower ranks than the uniform baseline. Similarly, the average assigned ranks for attention and FFN modules are also close to the uniform allocation. These observations suggest that the rank allocation strategy does not exhibit a systematic architectural preference toward specific layers or module types.

\textbf{Task-level analysis.} In contrast, we find that the allocation strategy varies more significantly across tasks than across network depth. As shown in Table~\ref{table: Average normalized assigned pruning ranks for different downstream tasks.}, some tasks consistently receive higher preserved ranks than the uniform baseline (e.g., SST-2, STS-B, QQP, and MNLI), whereas others are assigned lower ranks (e.g., CoLA, MRPC, QNLI, and RTE). This result suggests that the proposed strategy primarily adapts to the characteristics of the downstream task rather than to fixed architectural positions. It also supports our motivation that different fine-tuned models exhibit different robustness to rank pruning, making adaptive rank allocation preferable to a uniform pruning strategy.

\end{document}